**Route Planning for Last-Mile Deliveries Using Mobile Parcel Lockers: A Hybrid Q-Learning Network Approach**

**First and Corresponding Author**

**Yubin Liu**
Department of Civil and Environmental Engineering
Imperial College London, London, UK, SW7 2AZ
Email: y.liu20@imperial.ac.uk

**Co-authors List**

**Qiming Ye**
Department of Civil and Environmental Engineering
Imperial College London, London, UK, SW7 2AZ
Email: qiming.ye18@imperial.ac.uk

**Jose Escribano-Macias**
Department of Civil and Environmental Engineering
Imperial College London, London, UK, SW7 2AZ
Email: jose.escribano-macias11@imperial.ac.uk

**Yuxiang Feng**
Department of Civil and Environmental Engineering
Imperial College London, London, UK, SW7 2AZ
Email: y.feng19@imperial.ac.uk

**Eduardo Candela**
Department of Civil and Environmental Engineering
Imperial College London, London, UK, SW7 2AZ
Email: e.candela-garza19@imperial.ac.uk

**Panagiotis Angeloudis**
Department of Civil and Environmental Engineering
Imperial College London, London, UK, SW7 2AZ
Email: p.angeloudis@imperial.ac.uk

**ABSTRACT**

Mobile parcel lockers (MPLs) have been recently proposed by logistics operators as a technology that could help reduce traffic congestion and operational costs in urban freight distribution. Given their ability to relocate throughout their area of deployment, they hold the potential to improve customer accessibility and convenience. In this study, we formulate the Mobile Parcel Locker Problem (MPLP), a special case of the Location-Routing Problem (LRP) which determines the optimal stopover location for MPLs throughout the day and plans corresponding delivery routes. A Hybrid Q-Learning-Network-based Method (HQM) is developed to resolve the computational complexity of the resulting large problem instances while escaping local optima. In addition, the HQM is integrated with global and local search mechanisms to resolve the dilemma of exploration and exploitation faced by classic reinforcement learning (RL) methods. We examine the performance of HQM under different problem sizes (up to 200 nodes) and benchmarked it against the exact approach and Genetic Algorithm (GA). Our results indicate that HQM achieves better optimisation performance with shorter computation time than the exact approach solved by the Gurobi solver in large problem instances. Additionally, the average reward obtained by HQM is 1.96 times greater than GA's, which demonstrates that HQM has a better optimisation ability. Further, we identify critical factors that contribute to fleet size requirements, travel distances, and service delays. Our findings outline that the efficiency of MPLs is mainly contingent on the length of time windows and the deployment of MPL stopovers. Finally, we highlight managerial implications based on parametric analysis to provide guidance for logistics operators in the context of efficient last-mile distribution operations.

## 1. Introduction

The recent growth in e-commerce and online shopping transactions has prompted a considerable growth in demand for fast and timely parcel deliveries and underpins a significant increase in last-mile freight activities in urban areas, which have been shown to have a knock-on effect on traffic congestion, as well as on the demand for parking spaces (Savelsbergh and Van Woensel, 2016).

One of the solutions that have been recently adopted by logistics operators has been the deployment of stationary parcel lockers (SPLs), which enable unattended deliveries, simplifying delivery tours, and therefore reducing fleet size requirements (Deutsch and Golany, 2018). The service radius of such facilities, however, is limited by the walking distance of customers (Iwan et al., 2016). As a result, a very dense SPL network would have to be set up for such a scheme to be effective, requiring considerable capital investment and significant maintenance costs.

The emergence of increasingly capable electric and autonomous vehicle technologies has prompted the proposition of a new concept, namely mobile parcel lockers (MPLs), that has the potential to address some of the above limitations (Figure 1). MPL fleets could therefore respond to changing customer demand patterns, reduce facility number requirements, and increase accessibility for customers (J. Li et al., 2021). MPLs offer three potential advantages in urban parcel deliveries compared to traditional delivery vehicles. Firstly, MPLs are more flexible since they are not restricted to operating in highly congested areas (J. Li et al., 2021). Attempts will be made to facilitate MPL travel on pavements or bicycle lanes at slower speeds without occupying the main road that reducing traffic congestion. Secondly, MPLs can park in small curb side spaces, pavements, or public spaces outside buildings because of their small size, to increase customer convenience and accessibility. Finally, the mobility of MPLs enables them to provide multiple pick-up locations for customers, which has the potential to reduce repeated deliveries due to customer inconvenience. E-commerce operators have launched pilot studies to stimulate the broader application of MPLs in different regions, such as JD.com, Cainiao in China, and Google Mobile Delivery Platform in the US (Myllymaki, 2016). As another example, French car producer Renault introduced an autonomous driving platform which can load a parcel locker (EZ-Pro) that autonomously transfers packages towards customers (Joerss et al., 2016).

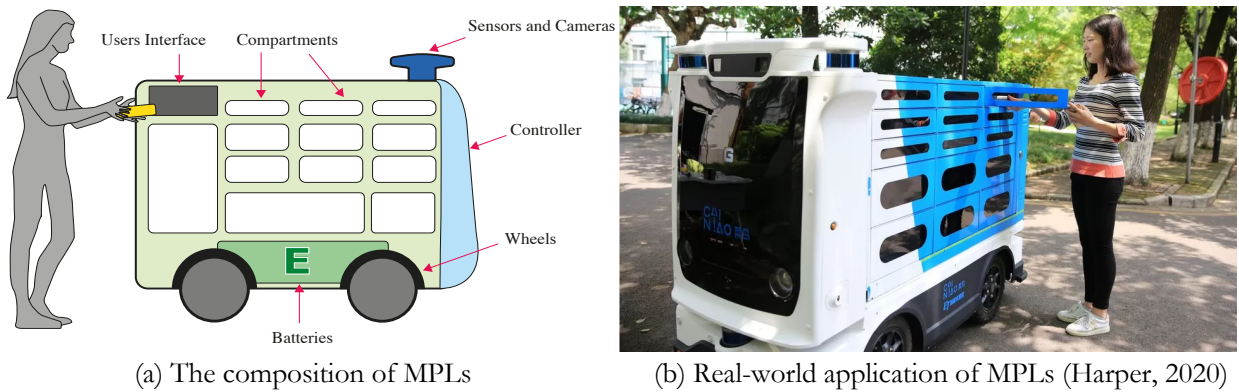

(a) The composition of MPLs      (b) Real-world application of MPLs (Harper, 2020)

**Fig. 1** Illustration of MPLs

Several studies have considered the prospect of applying MPLs in last-mile deliveries, but without fully exploiting the ability of this concept to dynamically adapt to varying levels of customer demand. For instance, Orenstein et al. (2019) proposes a model of delivering parcels from a depot to multiple service point using MPLs, in which recipients can choose more than one possible pick-up location. However, the author only considers a single period of the delivery round with a deterministic demand pattern. Wang et al. (2020) introduce MPLs to supplement fixed lockers, while aggregating demands from multiple delivery providers to achieve joint distribution. Nevertheless, the model did not consider customers' spatial distributions and the dynamic arrival of delivery requests. J. Li et al., (2021) proposes a two-echelon delivery network in which MPLs serve as an intermediate layer between couriers and depots, with the assumption, however, that customer locations and demands are static and deterministic. A more flexible version of this problem is studied by Schwerdfeger and Boysen (2020) with their Mobile Parcel Locker Problem (MPLP), where MPLs dynamically change their location in accordance with customer drop-off time/location preferences, and the objective of maximising end-user convenience. The limitation of the model is that the resupply of lockers with parcels is not considered, and the static characteristic of the model formulation make it difficult to capture such dynamic case.

Given the operational shortcomings of the above studies, this paper proposes a new parcel delivery framework that considers changing customers' demand patterns, parcel resupply, and identifies the critical factors contributing to MPLs' efficient operations. Figure 2 provides the schematic operation of MPLs. As a primary feature of our research, we use MPLs to dynamically transfer parcels between depots and available parking spaces, with special consideration for parking space availability and the drop-off location/timing requests that have been submitted by the customers. We assume customers provide multiple time/location combinations within the planning horizontal such that operators can synergistically determine the most appropriate delivery schemes. MPLs are then expected to deliver parcels to customers before returning to the depot so that MPLs can reload the next batch of parcels. Beyond the novel parcel delivery framework proposed above, we consider the dynamic planning technique that can resolve the time window conflict and delivery route adjustment, and we identify it as the critical difference between our model and the aforementioned MPL-related studies.

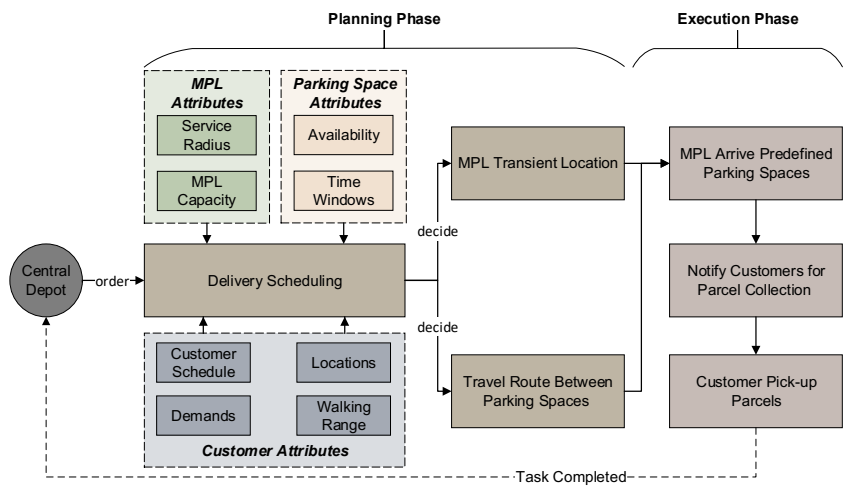

**Fig. 2** Illustration of schematic operation of MPLs

The solution approaches that have been previously used to solve MPLP include exact methods (Schwerdfeger and Boysen, 2020), traditional heuristics (Li et al., 2021), and meta-heuristics (Wang et al., 2020). However, the capability of the exact method is limited by the NP-hardness of MPLP, while heuristic algorithms are vulnerable to the presence of local optima. Reinforcement learning (RL) methods (e.g., Pointer networks, Attention mechanisms) have been previously used in Vehicle Routing Problems (VRPs), which provides the potential to be adopted to solve large MPLP instances. However, these learning techniques are highly dependent on the quality and availability of training samples and their pre-processing, causing low sampling efficiency (Sutton and Barto, 2018). To tackle the weakness of the above techniques, we propose a Hybrid Q-Learning-Network-based Method (HQM) with novel encoder and attention-based decoder to improve solution quality given its ability to exploit a wide spectrum of individual behaviour (Sutton and Barto, 2018). Further, the global and local search mechanisms are embedded within the HQM to improve local optima dilemma encountered by classic heuristics.

This study aims to develop an HQM that integrates global and local search mechanisms to solve MPLP considering customer demand patterns and time windows constraints. Concretely, the objectives of this paper are:

- **Model establishment:** establish a mixed integer programming model for MPLP that considers changing customers' demand patterns and parcel resupply.
- **Dynamic case modelling:** define dynamic route adjustment strategies to resolve time windows conflicts and on-demand MPLs allocation with varied conflicted scenarios.
- **Solution design (HQM):** establish a novel encoder & attention-based decoder framework of HQM for improving the training sample dependency and the exploration-exploitation dilemma faced by classic RL methods.
- **Performance measurement and parametric analysis:** evaluate the performance of HQM comparing to an exact approach and a meta-heuristic algorithm (GA) and define the effect of critical factors against fleet size, MPLs' travel distance, and service delay. Providing managerial and actionable insights for logistics professions based on parametric analysis.

In achieving these objectives, this paper contributes to the existing literature in the following respects:

- **Theoretical contribution:** to the best of the authors' knowledge, this research is the first attempt to apply an RL-based approach to solve MPLP. Our analysis reveals that HQM with improved sampling efficiency has better optimisation ability than the exact approach and GA in large MPLP instances regardless of the network type. We therefore aspire that it will serve as the foundation for future studies in the field of location-routing optimisation or VRPs in last-mile deliveries.
- **Algorithmic contribution:** our novel encoder & attention-based decoder framework of HQM is superior to other existing RL-based frameworks that use coordinates for training input as it can handle more decision parameters (as these parameters can be infinitely extended to the encoder structure) to resolve more complex multi-decision MPLPs and VRPs. The elaborate description of the encoder & attention-based decoder framework will be presented in section 4.3.1 & 4.3.3.
- **Modelling contribution:** we consider the dynamic planning scenario in our model where two route adjustment strategies are developed within the HQM framework to dynamically resolve the time windows conflicts caused by service delay and traffic conditions, which has the potential to be adopted to other RL-based methods in solving time-windows conflicts of VRPs.
- **Managerial contribution:** Based on parametric analysis of key factors that influence MPL-based parcel deliveries, we provide managerial insight (Section 5.4.4) from the perspective of logistics operators to stimulate broader adoption of autonomous last-mile delivery modes.

The remainder of the paper is structured as follows. Section 2 provides a brief review of major modelling approaches and algorithms for MPLP deployments. Section 3 describes the problem definition and formulation of a mathematical model for the MPLP. Section 4 presents our proposed solution approach for the MPLP instances. Section 5 presents the results of our experiments and evaluates the performance based on different aspects (e.g., final reward, convergence, computing time), compared with the Gurobi solver and GA-based alternative, used as the baseline. We further identify key influence factors (time windows, network sizes, capacity, etc) from both MPLs and customers' perspectives towards different cost units of the objective function. Additionally, we provide a managerial implication for future MPL operations based on the parametric analysis of different influencing factors. Finally, the conclusion section provides a summary of findings and highlights potential applications of our MPLP solution.

## 2. Literature review

This paper belongs to an emerging stream of research that focuses on the utilisation of autonomous vehicle technologies (MPLs) to Location-Routing Problems (LRPs) within the context of urban freight distribution. Therefore, we include the following fields related to our research for the purpose of this review: (i) LRPs, (ii) MPLP, and (iii) VRPs with RL approaches.

### 2.1 Location-Routing Problems (LRPs)

LRPs consist of two sub-problems: (1) determining the locations of various facilities, such as central depots and transshipment centers, and (2) assigning tasks and determining their execution sequences to meet customer demands, which is modeled as a VRP. In this context, we primarily focus on studies that address the solving sequence for LRPs, as this helps to define the solution structure for our model. The sequence for solving LRPs can generally be classified into sequential, iterative, and hierarchical methods (Nagy and Salhi, 2007).

Sequential methods follow a specific order to solve LRPs: either by first determining optimal locations for facilities and then determining routes based on these locations, or vice versa (Perboli et al., 2011). However, this approach ignores the information interaction between the location selection phase and the routing planning phase (Zhao et al., 2018). To address this issue, the solution optimality of sequential methods requires a comprehensive enumeration of all potential demand patterns in the model, and each demand must be met within the service radius to compromise the less interaction between the location selection phase and routing planning phase (Salhi and Rand, 1989; Enthoven et al., 2020).

Iterative approaches, which plan facility locations and vehicle routing simultaneously, address the issue of insufficient exploitation of information between the two sub-problems (Escobar et al., 2013; Jiang et al., 2019). Nevertheless, the capability of iterative approaches is limited by the increasing computational complexity of large problem instances and their reliance on accurate route estimation (Kaewploy and Sindhuchao, 2017; Prins et al., 2007).

Hierarchical methods solve the facility location problem as the main problem and the routing problem as the subordinate problem (Albareda-Sambola et al., 2005; Escobar et al., 2013; Melechovský et al., 2005). However, these methods rely on accurate route length estimation and may become stuck in local optima if the routing length is not accurately predicted in the location phase (Nagy and Salhi, 2007).

In our case, customer demand changes over time, and the location of parking spaces affects the location where customers can collect their parcels at each stage, whereas we can fulfil customers if they are within the service range of an MPL in one period. To the best of the authors' knowledge, this characteristic has not been defined as a dynamic facility location problem. Therefore, considering that we have enumerated all demand patterns for each customer within the service radius and each customer will be served at a predefined parking space during the planning horizon, we propose using a sequential approach to reduce computing complexity and route estimation inaccuracy that iterative and hierarchical methods faced in large problem instances.

The NP-hard nature of LRPs makes exact solutions inefficient and impractical for large problem instances. Therefore, heuristic (Duhamel et al., 2010) and meta-heuristic approaches (Yu et al., 2010) are commonly used to solve intermediate and large LRPs because they can find near-optimal solutions with less computation time. These heuristics methods include heuristic lagrangian relaxation (Yang et al., 2022), adaptive large neighbourhood search (ALNS) (Yu et al., 2021), ALNS integrated with iterative greedy algorithm and Clarke & Wright saving algorithm (Guo et al., 2022), GA-Simulated Annealing algorithm(Yu et al., 2022a), and NSGA-II (Heidari et al., 2022). The above studies have revealed that the effectiveness of (meta-)heuristics on improving solution quality for LRP and VRP. As a complement for the proposed RL method, we design a heuristic mechanism (global and local search) within HQM to compromise the computing complexity and solution quality in large problem instances.

We summarise classic studies based on the category of the solving sequence (Table 1). The main limitations of these studies include: (i) a lack of exploitation of the interaction between the two subproblems, leading to a significant deviation from the theoretically achievable optimal solution; (ii) a lack of consideration for customer uncertainty, such as changing demands, time slots, and locations, which may result in missed deliveries; and (iii) the potential for splitting customer demand to reduce total route length. In our case, limitations (i) and (iii) are addressed by enumerating all possible pick-up locations/stopovers of customers before the route planning phase, ensuring that all potential pick-up locations of each customer can be considered in the delivery scheme. The demand within the same parking space can be met by different MPLs to minimise the fleet size. To address limitation (ii), we develop different route adjustment strategies to reduce service delays caused by time window conflicts.

**Table 1.** Summary table of classic LRP studies

| Category | Reference | Main problem/Method | Network type/size | Objective | Key constraints and features |
|---|---|---|---|---|---|
| Sequential methods | Enthoven et al. (2020) | 2E-VRP-CO/MIP, ALNS. | CDPs/1 depot, up to 200 customers, 50 lockers, 10 satellites. | Minimise the traveling costs and the connection costs. | Fleet size, flow conservation constraints, covering options, vehicle capacity. |
| | Perboli et al. (2011) | CVRP/ MIP formulation and valid inequalities. | 2E-DP/Up to 50 customers, 5 satellites. | Minimises the traveling and handling operations costs. | Locker capacity, arc flow of each echelon, demand fulfilment. |
| | Zhao et al. (2018) | CLRP/Cooperative approximation heuristic. | 2E-attended-and-unattended-CDPs/Up to 200 customers, 20 depots | Minimise vehicle routing costs. | Heterogeneous fleet, carbon emissions, demand variations. |
| Iterative methods | Escobar et al. (2013) | CLRP/Granular tabu search-based heuristics. | DPs/Up to 20 depots, 200 customers. | Minimise depot costs and the traveling costs of vehicles. | Each cluster is assigned to exactly one depot, depot capacity. |
| | Jiang et al. (2019) | TSP-FLP/Iterated local Search. | CDPs/1 single route, 1 depot, 81 customers, 5 lockers. | Minimise the delivery cost and carbon emission. | Demand fulfilment, flow balance, single delivery route. |
| | Kaewploy and Sindhuchao (2017) | CLRP/Tabu search. | DPs/16 depots, 5 routes, 5 vehicles. | Minimise transportation costs and depreciation costs. | Depot and vehicle capacities, flow balance, prevention of sub-tours, demand variations. |
| | Prins et al. (2007) | CLRP/Lagrangean relaxation, granular tabu search | DPs/Up to 10 depots, 200 customers. | Minimise depot fixed cost and traveling costs. | Depot and vehicle capacities, subtour elimination, flow balance, demand fulfilment. |
| Hierarchical methods | Albareda-Sambola et al. (2005) | LRP/LP, Tabu search | DPs/Up to 5 depots, 30 customers. | Minimise depot fixed cost and traveling costs. | Depot and vehicle capacities, flow balance, demand fulfilment. |
| | Duhamel et al. (2010) | VRPRB/Path relinking algorithm. | DPs/Up to 5 depots, 200 customers. | Minimise routing cost and the difference between the largest and smallest route cost. | Deliveries cannot be split, each route starts and ends at the depot, vehicle capacity. |
| | Melechovský et al. (2005) | LRIP/Non-linear MIP, MOGWO, NSGA-II | Multi-echelon DPs/1 depot, 3 DC, 3 BC, 300 customers. | Minimise fixed, inventory, transportation, and emission costs. | Demand fulfilment, flow balance, depot and vehicle capacities, time-windows, inventory level, order value. |
| | Nadizadeh and Hosseini Nasab (2014) | CLRP-FD/Stochastic simulation, local search | DPs/Up to 5 depots, 100 customers. | Minimise fixed costs of depots and vehicles, traveling and opportunity lost costs | Fuzzy demands, depot and vehicle capacities, deliveries cannot be split, travel distance limitation. |
| | Yu et al. (2010) | CLRP/SA | DPs, vehicle without return to depot/Up 4 depots, 318 customers. | Minimise depot fixed and operational costs, vehicle traveling costs. | Demand fulfilment, vehicle and flow capacities, time-windows |
| | Zare Mehrjerdi and Nadizadeh (2013) | CLRP-FD/Greedy clustering method, Stochastic simulation | DPs/Up to 7 depots, 100 customers | Minimise fixed costs of depots and vehicles, traveling costs | Fuzzy demands, depot and vehicle capacities, deliveries cannot be split, route quantity. |
| | Zhang et al. (2019) | MOVRPFlexTW/ACO, pareto optimality | DPs/Solomon benchmarks | Minimise routing costs. Maximise customer satisfaction. | Vehicle capacity, time windows, demand fulfilment. |

**Notes:** 1) ALNS=Adaptive large neighbourhood search; 2) ACO=Ant colony optimisation; 3) BC= Backup centre; 4) CDP=Collection and delivery problems; 5) CLRP-FD=dynamic capacitated location-routing problem with fuzzy demands; 6) DC=Distribution centre; 7) DP=Delivery problems; 8) 2E-DP=2-Echelon delivery problems; 9) LRIP =Location, routing and inventory problem; 10) MOGWO=Multi-objective gray wolf optimisation algorithms; 11) MOVRPFlexTW= Multi-objective VRP with flexible time windows; 12) NSGA-II=Non-Dominated Sorting Genetic algorithm; 13) VRPRB=VRP with route balancing;

*2.2 Mobile Parcel Locker Problem (MPLP)*

The MPLP is a variant of the LRP. While traditional LRPs consider the location selection problem as the main issue, existing MPLP studies focus on route optimisation for MPLs (referred to as the main problem). To solve MPLP, decision-makers must first define optimal stopovers for lockers and then plan delivery routes for MPLs accordingly (following the sequential solving structure mentioned in Section 2.1). Schwerdfeger and Boysen (2020) introduced the first MPLP as a mixed-integer programming model that considers customer coverage and minimises the MPL fleet size by determining MPL stopovers and their corresponding routes.

While their exact solution approach demonstrates significant improvements compared to stationary locker systems, several key limitations arise: (i) it does not consider the dynamic scenario (e.g., parcel resupply) and stochastic factors, such as service delays, which are present in real-world operations and can affect the dynamic adjustment of scheduling strategies; (ii) it does not track the operational details of a single MPL, such as starting and ending times and duration at parking spaces, which limits in-depth route optimisation for a single MPL and leads to a larger gap from the optimal solution; (iii) from a computational perspective, the efficiency of the proposed exact solution approach is significantly lower (computing time ≈ 60min) than most evolutionary algorithms in large networks (nodes≥100).

Our study differs from Schwerdfeger and Boysen (2020) in several ways. We address the aforementioned first limitation by incorporating service delay and travel distance into the objective function and implementing route adjustment strategies to address latency considering parcel resupply. To address the second limitation, we design a state-based structure for solution extraction that allows the delivery scheme of each MPL to be interpreted and revised by the proposed HQM. To address the final limitation, we integrate HQM with global and local search mechanisms to achieve a better balance between solution quality and computational efficiency compared to traditional exact and heuristic approaches.

Orenstein et al. (2019) proposed a flexible parcel delivery problem that allows customers to provide multiple delivery locations to couriers. However, this model does not account for the scenario that customers may change their preferred destinations if initial delivery attempts fail. This is a significant consideration, as it is estimated that 42% of orders delivered to consumers without specific delivery times will result in repeated delivery attempts (Allen et al., 2017). To address this issue, our proposed model enumerates all potential customer stopovers and considers the overlap of the corresponding time windows when formulating the delivery scenario to ensure that all demands can be met at specific parking spaces.

Wang et al. (2020) integrated the LRPs with MPLs and established a non-linear integer programming model to minimise operating costs. The results show that the policy that considers demand aggregation can significantly reduce delivery times while a scheme without demand aggregation would reduce the number of vehicles required. Hence, we consider demand aggregation in our scenario to reduce fleet size and travel distances.

Finally, J. Li et al. (2021) introduced a two-echelon MPLP, in which locker travel to couriers in the field and transfer parcels between the couriers and the depot, while couriers perform the final delivery without making repeated returns to the depot. A hybrid Clark and Wright heuristic was developed to solve the mixed-integer programming model. The common weakness of the model introduced by J. Li et al. (2021) and Wang et al. (2020) is that all customer demand patterns are assumed to be static in contrast to the stochastic nature of real-world customer requests, and the proposed method becomes computationally infeasible for larger problem instances.

Based on the reviewed literature on MPLP, we conclude that existing studies consider deterministic cases where demand patterns and delivery routes are static. This can lead to lower optimisation performance when faced with stochastic events such as demand changes and traffic congestion. Our approach addresses this issue through predicting the time consumption for predefined delivery routes before tasks are executed, allowing for predictions and route adjustments to be made by the proposed algorithm in each iteration. A summary of the key features of the aforementioned studies can be found in Table 2.

**Table 2.** Summary table for the MPLP studies

| Reference | Model/Approach | Network type/size | Objective | Key constraints |
|---|---|---|---|---|
| Schwerdfeger and Boysen (2020) | Dynamic FLP/ Domain reduction, MIPs, exact approach | CDPs/Up to 300 customers, 40 locations | Minimise fleet size | Time windows, locker capacity, service radius |
| Orenstein et al. (2019) | CVRP/Saving heuristics and tabu search, LNS | CDPs with alternative delivery points/Up to 50 service points | Minimise fixed and travel costs of lockers, and the penalty of undelivered parcels | Time windows, locker capacity, demand fulfilment, split deliveries are not allowed, fleet size at SP |
| Wang et al. (2020) | CVRP with demand aggregation/MIPs, Multi-layer GA | CDPs with alternative collection points/32 nodes | Minimise transportation and maintenance costs | Time windows, locker capacity, service range, demand fulfilment, fleet size at SP |
| J. Li et al. (2021) | LRP/Hybrid Clark and Wright heuristic | CDPs with 2-Echelon/Up to 4 depots, 200 customers | Minimise depot rental, locker deployment and the operational costs | Locker capacity, service radius, demand fulfilment, visiting times at SP |

**Notes:** 1) FLP=Facility location problem; 2) MIPs=Mixed integer programming; 3) CVRP=Capacitated vehicle routing problem; 4) LNS= Large neighbourhood search; 5) CDPs= Collection and delivery problems; 6) SP=Service point

*2.3 Applications of Reinforcement Learning in Vehicle Routing Problem*

Research in the field of RL has demonstrated its potential for effectively solving large, time-sensitive VRPs by leveraging environmental cues to find better solutions. (Bello et al., 2017; Cappart et al., 2020; Nazari et al., 2018). We therefore review the features of different RL structures and the key design elements of RL frameworks, rather than providing details on implementation procedures, to provide insight into the design of the HQM structure.

RL-based approaches that have been used to solve VRPs can be divided into two categories: i) end-to-end methods (e.g., Pointer network, Ptr-Net) and ii) the use of deep RL (DRL) to improve iterative search algorithms. End-to-end methods provide a scalable and efficient framework for optimising deep neural networks, but they tend to produce lower quality solutions for medium to large problem instances compared to other combinatorial optimisation solvers such as LKH3, Google OR Tools, and Gurobi. (Wang and Tang, 2021).

DRL-based iterative search frameworks use neural networks to learn and choose heuristic rules within heuristic algorithms to improve the speed of generating new solutions for classical heuristics in large problem instances. While these frameworks offer better solution quality compared to end-to-end methods, they sacrifice generalisation ability in solving similar type of problems. Hence, the trade-off between optimisation capability and generalisation ability in different problem instances should be considered due to the nature of RL-based methods (Sutton and Barto, 2018).

Table 3 summarises the typical studies of each RL technique mentioned above. Based on the analysis of reviewed literature, we notice that the solution quality of RL techniques will be affected by the performance of sampling, which is determined by the structure of encoder & decoder (Yu et al., 2019) or the state interpretation (Zhao et al., 2021). Hence, the encoder of our HQM is designed to contain a set of vectors that combine task allocation tuples and task execution sequence tuples, such that provides better initial solution for the main optimisation procedure/loop. Compared with the classic encoder that only uses the coordinates of nodes in traditional pointer networks (Deudon et al., 2018; Vaswani et al., 2017; Vinyals et al., 2017), the HQM's encoder contains more decision parameters without transforming the original coordinates, improving computing speed and solution quality while can be extended to formulate multiple decision variables accordingly.

Beyond the above processing in sampling phase, we develop a novel attention-based mechanism to decode the processed vector with superior solutions during the state interpretation and transition phase. The optimal strategies are obtained through continuous training with a trial-and-error mechanism. The training process of HQM follows an offline manner and does not rely on historical samples.

Compared with other online training DRL methods with on-policy or off-policy sampling, the offline training in HQM will significantly improve the sample efficiency since the offline data/sample will be generated based on the behaviour policy. To overcome the weakness of solution quality faced by most off-line RL methods, we introduce $\varepsilon - \text{greedy}$, and global and local search mechanisms to balance the exploitation-exploration dilemma. The detailed description of the HQM will be introduced in Section 4.3.

**Table 3.** Classification of RL methods in solving VRP

| Classification | Study | Model & Training Method | Problem Instance & Performance |
|---|---|---|---|
| End-to-end Method (Pointer Network) | Vinyals et al. (2017) | Ptr-Net; supervised training | 30-TSP: near optimal and better than heuristics algorithm; 40, 50-TSP: significant gap towards optimality |
| | Bello et al. (2017) | Ptr-Net; REINFORCE & Critic baseline | 50-TSP: solution better than Vinyals et al. (2017). 100-TSP: near-optimal solved by Concorde. 200-Knapsack: optimal solution. |
| | Nazari et al. (2018) | Ptr-Net; REINFORCE & Critic baseline | 100-TSP: save training time around 60% compared with (Bello et al., 2017; Deudon et al., 2018; Nazari et al., 2018; Vinyals et al., 2017). 100-CVRP: better than heuristics algorithm. |
| | Deudon et al. (2018) | Transformer attention REINFORCE & Critic baseline | 20, 50-TSP: solution obtained better than Bello et al. (2017). 100-TSP: result similar to Bello et al. (2017). |
| | Vaswani et al. (2017) | Transformer attention REINFORCE & Rollout baseline | 100-TSP: solution obtained better than previous research (Bello et al., 2017; Deudon et al., 2018; Nazari et al., 2018; Vinyals et al., 2017) . 100-CVRP, 100-SDVRP, 100-OPVRP, 100-PCTSP, SPCTSP: near optimal solved by Gurobi, and better than heuristics algorithms. |
| | Kool et al. (2019) | Graph pointer network; HRL | 20, 50, 250, 500, 1000-TSP: solution obtained better than Bello et al. (2017). 20-TSPTW: solution obtained better than OR-Tools、and ant colony optimisation. |
| | Ma et al. (2019) | Ptr-Net; REINFORCE & Critic baseline & Decompose policy/parameter migration | 100, 150, 200, 500-Multi objective TSP: better than MOEA/D, NSGA-II, and MOGLS. |
| DRL-based Iterative Method | Chen and Tian (2019) | Graph attention & PPO | 100-CVPR: solution obtained better than Vaswani et al. (2017). 100-CVPRTW: better than multi-heuristics algorithms. |
| | Gao et al. (2020) | Transformer attention & REINFORCE | 20, 50, 100-CVRP: solution obtained better than Nazari et al. (2018), Vaswani et al. (2017), OR Tools, and LKH3. Computing speed better than LKH3. |

**Notes:** 1) SDVRP=Split delivery VRP; 2) OPVRP=Order picking VRP; 3) PCTSP= Prize collecting TSP; 4) Stochastic prize collecting TSP; 5) MOEA/D=Multi-objective evolutionary algorithm based on decomposition; 6) NSGA-II=Non-dominated sorting genetic algorithm II; 7) MOGLS=Multi-objective genetic local search

## 3. Problem statement

We propose an MPLP model for parcel deliveries considering customers' changing locations and time windows. A set of customer nodes $N$ serves as the recipient of parcels from a group of MPLs. For each customer $n \in N$, customer $n$ creates a set of locations $K_n$ during the planning horizon. Each location $k$ ($k \in \{1, 2, \cdots, \|k_n\|\}$) that a customer visited within the location set $K_n$ is defined by a time interval $[E_{nk}, L_{nk}]$ and a corresponding position $[X_{nk}, Y_{nk}]$. To better fit the real-world operations, we introduce customer's maximum walking range $\rho_n$ to investigate how it affects the delivery system (will be introduced in Section 5.4).

A set of MPLs is denoted by $M$, which MPLs can change their locations to facilitate parcel pickups by the customer set $N$. The service radius of each MPL $m$ ($m \in M$) is denoted as $r_m$, and the average travel speed is $v$. We define a set of parking spaces $I$ where locker $m$ can potentially park to serve customers. Each parking space $i \in I$ is represented by a position $[X_i, Y_i]$, and the available parking time windows $[E_i, L_i]$. We use $SV_i$ to denote the time consumption of parking space $i$ for serving each customer. Therefore, for each parking space $i$, the parking time windows $[E_i, L_i]$ can be divided into $\|U_i\|$ ($\|U_i\| = |L_i - E_i/SV_i|$) sub-intervals, where $U_i$ represents a set of sub-interval time spans in parking space $i$. Note that $\|U_i\|$ needs to be rounded up if it is not an integer such that ensure the length of the original available parking time windows $[E_i, L_i]$ is fully considered. We introduce $[e_{ia}, l_{ia}]$ to represent the $a^{th}$ ($a = 1, \ldots, \|U_i\|$) sub-interval within $U_i$ in parking space $i$ (i.e., $[e_{ia}, l_{ia}] \in [E_i, L_i]$). Let $U$ ($U = \{U_1, U_2, \cdots, U_i\}$) to represent a set of all sub-intervals for all parking space $i$.

We define a task $O_{ima}$ as the movement of MPL $m$ to a specific parking space $i$, fulfilling the demand within the $a^{th}$ sub-interval $[e_{ia}, l_{ia}]$ of parking space $i$ (i.e., Each sub-interval time span corresponds to one single task). Let $O$ denotes a set of all tasks to be performed ($O_{ima} \in O$). By obtaining a set of tasks to be fulfilled at different sub-intervals of all parking spaces $i \in I$, the tasks are assigned to MPLs to perform while considering the execution sequence. The task execution sequence of MPLs, therefore, formatting the VRP. Our proposed MPLP makes the following decisions in sequence:

1) The optimal location of a set of parking spaces $I$, such that customers within the service radius can access MPLs for parcel collection.
2) The task assignment scheme for all determined tasks (i.e., by which task to be performed by which MPL), minimising the MPLs deployed.
3) The task execution sequence for all MPLs (i.e., the route of MPLs visiting different parking spaces to perform tasks), considering customers' time-windows, available parking time, and shorter route length.

We do not plan the route between customers and MPLs since each customer will only access the nearest MPL (this only composes a direct shortest path between MPLs and customer locations).

As a result of MPLs' delay, customers might not be served by MPLs on time. In this case, an additional travel distance arises when modifying predefined delivery routes. This route modification is designed to accommodate the undelivered parcel (the corresponding route adjustment algorithms will be introduced in *Route adjustment strategy and route generation algorithm* section).

Figure 3 illustrates an example of MPLP with three customers and one MPL. The available parking spaces for MPLs are denoted by solid triangles from A to D, and the corresponding available parking time windows are represented by ovals. The three customers are denoted by the solid circle with different numbers from 1 to 3, and their available pickup times at different locations are represented as rectangles. Customers may change location throughout the day as denoted by the dotted arrows. For example, Customer 1 departs from the location covered by Parking Space A to the place covered by Parking Space B, and finally reaches Parking

Space D. Customer 2 travels from the location covered by Parking Space C to the destination covered by Parking Space B. Analogously, Customer 3 travels from the location covered by Parking Space C to the destination covered by Parking Space A. The optimal solution in this example is to deploy one MPL, which leaves the depot and visits Parking Space C and Parking Space D in sequence before returning to the start when completion (since the available parking time intervals satisfy both customer time windows and service buffer time).

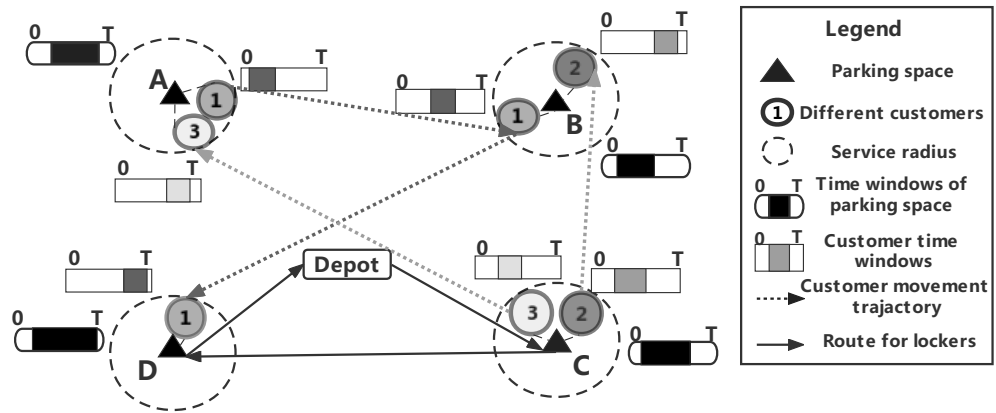

**Fig. 3** Instance of MPLP

*3.1 Model assumptions*

When developing our model, we made the following assumptions:

**Assumption 1:** It is assumed that the logistics operator will ask customers (via the mobile APP or message) if he/she would like to receive MPL service before parcel deliveries. If the customer accepts, they can collect parcels at one of the parking spaces based on their stopovers within a day. The logistics operator will dispatch the MPL to the predefined parking space to serve the customer according to the optimal dispatch strategy. Since the logistics operators only require the approximate location of the customer prior to the service instead of a real-time and precise trajectory, the data privacy of customers can be well-protected. Other potential incentives of the logistics operators to overcome these concerns, e.g., reduced service fares and a better mechanism for protecting customers' privacy, is worth considering in practice.

**Assumption 2:** The time windows of a specific customer at different locations shall never overlap to fit real operations.

**Assumption 3:** We assume the maximum accepted walking ranges of customer $\rho_n$ is a constant which we set different values to evaluate its influence on the delivery system (see Section 5.4). However, $\rho_n$ can vary from different customers following a specific probability distribution if required. In addition, the available parking duration $[E_i, L_i]$ varies for different parking spaces due to the different regulations in the region.

**Assumption 4:** The service charge for parking spaces is not considered. We assume that parking spaces have infinite capacities to accommodate multiple MPLs occupying a parking space. However, the capacity of parking spaces can be adjusted and limited according to real operation.

**Assumption 5:** Each locker will load parcels at the beginning of the planning horizon and will drive back to the depot once the MPL tasks in the current delivery round are completed.

**Assumption 6:** Since parcels are standardised in their maximum dimension and weight by logistics providers, we assume that all parcels fit any compartment of the MPL.

*3.2 Mathematical model for MPLP*

Following the above problem statement, the proposed MPLP model consists of three parts, as shown in Figure 4. The first phase is the location selection, in which the optimal locations for parking spaces are determined based on the spatial distribution of customers. In this phase, all potential stopovers of customers are enumerated and the K-means clustering method is applied to identify the coordinates of optimal parking spaces and the customers' location-parking space pairings, which are used as inputs for the route planning phase. The second phase, route planning, aims to determine delivery routes for MPLs to minimise the MPL fleet size, total distance travelled, and service delays while satisfying all customers and parking time window constraints. To achieve this goal, we first generate the initial feasible solution, which consists of a task assignment scheme and task execution sequence. The generated initial solution will be inputted into the HQM module, where the initial feasible solution will be encapsulated as a state to make it interpretable by the main components of HQM. The main body of HQM includes several crucial optimising components to obtain better state/solution through iteration. The detailed procedure of the HQM module will be introduced in Section 4. The notation and the mathematical model for the first two phases are as follows:

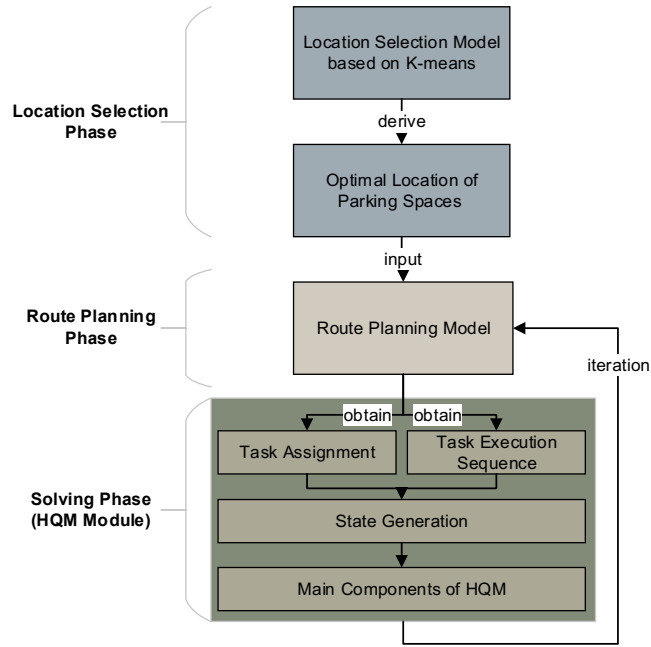

**Fig. 4** Illustration of the model relationship

**Indices:**

| | | | |
|---|---|---|---|
| $m$ | MPL $m$ ($m \in M$) | | |
| $n$ | Customer $n$ ($n \in N$) | | |
| $i, j$ | Two adjacent parking spaces ($i, j \in I$ in route planning phase, $i \in \{1, \cdots, P\}$ in location selection phase) | | |
| $k$ | The $k^{th}$ location of a single customer ($k \in K_n$) | | |
| $a$ | The $a^{th}$ task of a parking space $i$ ($a \in U_i$) | | |

**Sets:**

| | |
|---|---|
| $N$ | Set of customers |
| $M$ | Set of MPLs |
| $I$ | Set of parking space (excluding depot) |
| $V$ | Set of nodes within the network (including set $I$ and depot) |
| $K_n$ | Set of locations for a customer $n$ |
| $U_i$ | Set of sub-interval/tasks in parking space $i$ |
| $U$ | Set of all sub-intervals/tasks for all parking space |

**Parameters:**

**Auxiliary Variables:**

| | | | |
|---|---|---|---|
| $Q$ | The capacity of MPLs | $A_{ima}$ | The time when locker $m$ arrives at parking space $i$ to perform the $a^{th}$ task of parking space $i$ |
| $r_m$ | Service radius of MPL $m$ | $q_{ia}$ | The total demand to be fulfilled within the $a^{th}$ sub-interval of parking space $i$ |
| $F_m$ | Fixed cost of MPL $m$ | $[e_{ia}, l_{ia}]$ | The sub-interval within $[E_i, L_i]$ of the $a^{th}$ task at parking space $i$ ($a = 1, \ldots \|U_i\|$) |
| $v$ | Travel speed of MPLs | $d_{ij}$ | Distance from parking space $i$ to $j$ |
| $h_{nk}$ | Coordinates of the $k^{th}$ location of customer $n$ ($h_{nk} = [X_{nk}, Y_{nk}]$) | $d_{in}^k$ | Distance from the $k^{th}$ location of customer $n$ to parking space $i$ |
| $\rho_n$ | Maximum walking range of customer $n$ | $t_{ij}$ | Actual travel time from parking space $i$ to $j$ |
| $c_{ij}$ | Unit cost travelling from parking space $i$ to $j$ | **Decision Variables:** | |
| $\phi$ | A large indefinite number | $g_i$ | Float: coordinates of the parking space $i$ ($g_i = [X_i, Y_i]$) |
| $SV_i$ | Service time of parking space $i$ | $b_{nk}$ | Integer: the cluster index of $k^{th}$ location of customer $n$ (serve as true labels) |
| $\Psi$ | The maximum number of MPLs that can be used | $x_m$ | Boolean: 1, if locker $m$ is used for service; 0, otherwise |
| $P$ | The total number of parking spaces (clusters) needs to be determined (use for location selection phase) | $y_{ijm}$ | Boolean: 1, if locker $m$ travel to parking space $j$ from parking space $i$; 0, otherwise |
| | | $z_{ima}$ | Boolean: 1, if locker $m$ fulfils the $a^{th}$ task of parking space $i$; 0, otherwise |

### 3.2.1 Location model for parking spaces

Following the notation defined in Section 3.2, we establish a location selection model for parking spaces as follows (adapted from Andrew Ng (2000)):

$$b_{nk} = f(g_i) = \arg\min_{i \in \{1, \cdots P\}} \|h_{nk} - g_i\|^2 \qquad \forall n \in N, k \in L_n \tag{1}$$

where:

$$g_i = \frac{\sum_{n=1}^{N} \sum_{k}^{K_n} 1\{b_{nk} = i\} h_{nk}}{\sum_{n=1}^{N} \sum_{k}^{K_n} 1\{b_{nk} = i\}} \qquad \forall i \in \{1, \cdots P\} \tag{2}$$

$$g_i \in \mathbb{R}, b_{nk} \in \mathbb{N} \qquad \forall i \in \{1, \cdots P\}, n \in N, k \in L_n \tag{3}$$

The location model intends to partition all customers' location (data point) into $P$ parking spaces (clusters) in which each customer's location belongs to the parking space with the nearest mean. The objective of K-Means clustering is to determine the cluster index $b_{nk}$ (true label) for each customer's location $h_{nk}$ through minimising squared error (total intra-cluster variance) in Equation 1 (making this an unsupervised learning problem). Equation (2) represents the calculation for all centroids of parking spaces through customer location-parking space matching. Equation (3) defines the domains of the decision variables.

Since the K-means requires an explicit number of clustering centres $P$ (parking spaces) before implementation, we can determine the minimum value of $P$ by guarantying $d_{in}^k \leq \rho_n \leq r_m$ through trial and error.

### 3.2.2 Route planning model for MPLs

By implementing the location model, the obtained optimal location of parking spaces and the customer location-parking space pairing are used as a part of the input of the route planning model. This allows the routing model to calculate the distance between customers and parking spaces. The route planning model for MPLs is demonstrated as follows:

$$\text{Minimise } f\left(x_m, y_{ijm}, z_{ima}\right) = W_1 C_1 + W_2 C_2 + W_3 C_3 \tag{4}$$

$$C_1 = F_m \sum_{m=1}^{M} x_m \tag{5}$$

$$C_2 = \sum_{m=1}^{M} \sum_{i=0}^{V} \sum_{j=0}^{V} y_{ijm}\, d_{ij}\, c_{ij} \tag{6}$$

$$C_3 = \sum_{m=1}^{M} \sum_{i=1}^{I} \sum_{a}^{U_i} [\max(e_{ia} - A_{ima}, 0) + \max(A_{ima} - l_{ia}, 0)] \tag{7}$$

Subject to:

$$\sum_{m=1}^{M} x_m \leq \Psi \tag{8}$$

$$\sum_{i=1}^{I} \sum_{a=1}^{U_i} z_{ima} \cdot q_{ia} \leq Q \cdot x_m \qquad \forall m \in M \tag{9}$$

$$\sum_{m=1}^{M} z_{ima} = 1 \qquad\qquad \forall i \in I, a \in U_i \tag{10}$$

$$\sum_{i=1}^{I} y_{0im} = \sum_{m=1}^{M} x_m \tag{11}$$

$$\sum_{j=1}^{I} y_{j0m} = \sum_{m=1}^{M} x_m \tag{12}$$

$$A_{jma} = A_{ima} + SV_i + d_{ij}/v - \phi\left(1 - y_{ijm}\right) \quad i, j \in V; i \neq j; m \in M; a \in U_i \tag{13}$$

$$x_m, y_{ijm}, z_{ima} \in \{0, 1\} \qquad \forall m \in M; i, j \in V; a \in U_i \tag{14}$$

The objective function (4) minimises the fleet size, total travel distance and service delay, where $W_1$, $W_2$, and $W_3$ represent the weights of three cost units. Equation (5) to (7) denotes the MPLs' deployment cost, travel cost, and delay penalty respectively.

Constraint (8) denotes the total quantity of MPLs used should not exceed the predefined value. Constraints (9) represents the total demand fulfilled by locker $m$ should not exceed the maximum capacity of MPL. Constraint (10) ensures that each task within the parking space $i$ will be executed only once. Constraints (11) and (12) guarantee MPLs depart from a central depot and will return to the depot when the task is completed, and the number of routes from and back to the depot is equal to the MPL fleet size. Constraint (13) represents the time relationship of locker $m$ visit to parking space $j$ from parking space $i$. Constraint (14) defines the domains of the decision variables.

## 4. Methodology

The structure of our proposed solution approach is illustrated in Figure 5. Firstly, we utilise K-means clustering to determine the optimal location for parking spaces based on customer distribution and assign different customer locations to the nearest parking space, so that a customer can access MPLs at parking spaces at one of their potential locations. Since locating parking spaces through K-means clustering is not considered to be the main problem in our study, and a number of open-source packages have already been developed for realising K-means clustering, we will not provide detailed instructions for implementation. Related studies of applying K-means for location selection problems see Kumar and Somasundaram Kumanan (2012)and Pham et al. (2005).

Next, the task generator produces task set $O$ for MPL fleet. MPL $m$ are then allocated to perform each task $O_{ima}$ ($O_{ima} \in O$) within the $a^{th}$ sub-interval $[e_{ia}, l_{ia}]$ in parking space $i$ (as described in Section 3).

Then, the HQM algorithm is developed to determine the following set of decisions: i) the task assignment schemes (denoted as $x_1$) that define the task-locker assignment, and ii) the task execution sequence for each MPL (denoted as $x_2$). The length of $x_1$ and $x_2$ corresponds to the total number of tasks ( $\|O\|$) determined in the task generation phase. A state $s = (x_1, x_2)$, which consists of a $x_1, x_2$ pairing, represents a feasible solution of the proposed MPLP. A set of state $s$ forms an agent $\pi$. The reward $R$ of HQM is defined as the reciprocal of the objective function (4) since the HQM is set to maximise the reward. The action $A_t$ is defined as a decision behaviour in which the agent $\pi$ selects a policy to obtain a better solution in the next timestep $t$.

The HQM contains an encoder, the global and local search framework, a route generator, and an attention-based decoder. The encoder first produces an agent $\pi$ through a random generator and permutation pool. To speed up the searching process for the optimal solution, the global search mechanism based on the Q-value matrix and the local search mechanism based on single-state adjustments are used concurrently to generate the new state $s'$ for the agent $\pi$.

The updated state $s'$ will then be input into the route generator so that the delivery routes for MPLs can be developed. The decoder is used for calculating the reward $R$ of a state $s$ and decoding the best state $s^*$ such that obtains the optimal task assignment scheme $x_1^*$ and the tasks execution sequence $x_2^*$. The attention layer within the decoder is used to specify the selecting possibility of each element within $x_1$ and $x_2$ that is inherited by the next action $A_{t+1}$. The current best state $s^*$ will then be recorded, and the Q-value matrix will be updated accordingly. Once the convergence condition has been satisfied, the HQM outputs the best state $s^*$.

The encoder generates an agent $\pi$ that is input to the decoder, which then identifies optimal state $s^*$ for each agent $\pi$ according to the current knowledge of the problem. In this manner, the optimal task assignment scheme $x_1^*$ and the task execution sequence $x_2^*$ are determined. A description of the detailed structure and essential elements of the HQM will be provided in Section 4.3. The following section will describe the detailed mechanisms of the various components within the solution framework in sequential order.

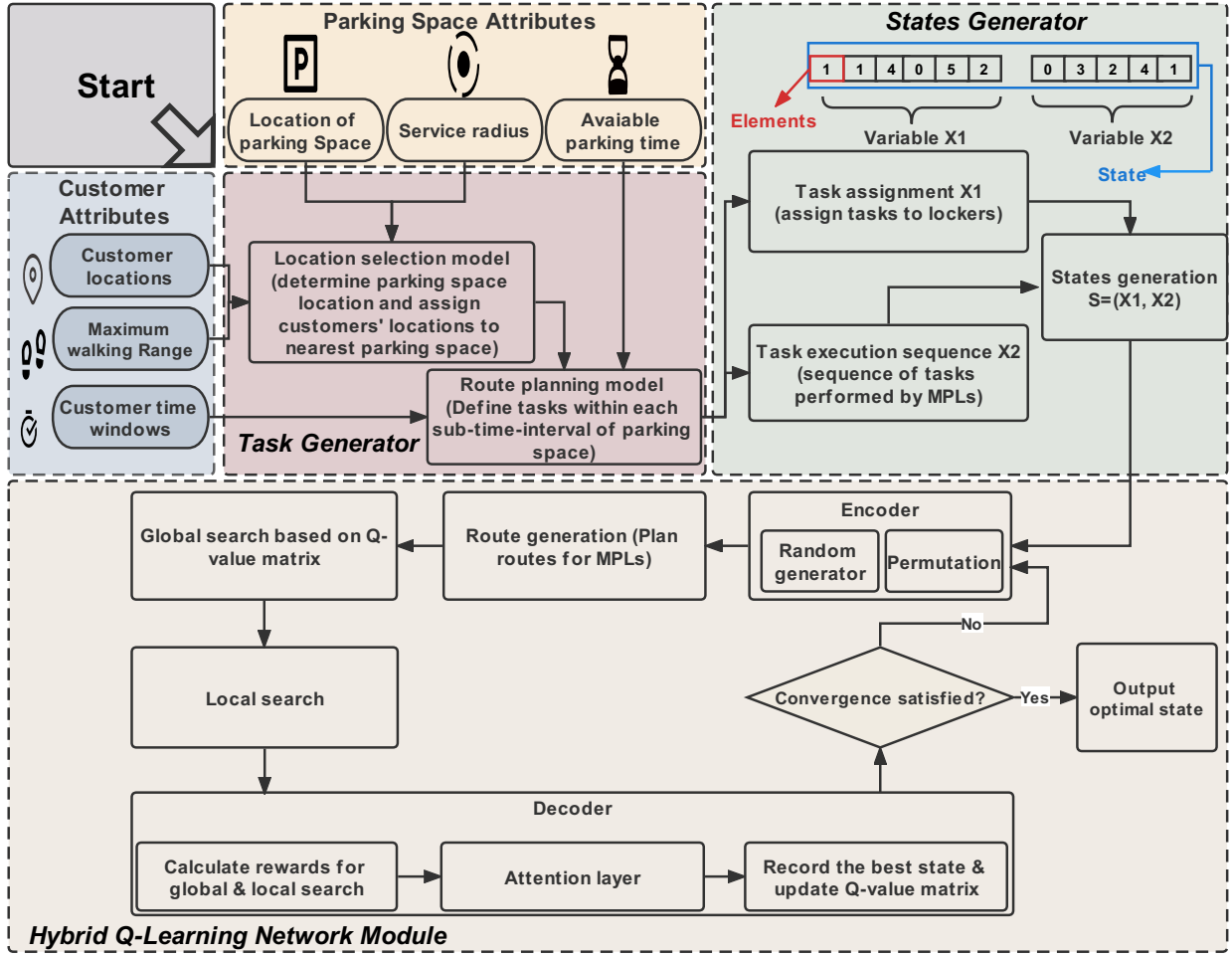

**Fig. 5** Solution framework of the model

*4.1 Task Generation*

Following the procedure presented in Figure 5, we first demonstrate the process of task generation. The objective of this phase is to identify a set of tasks that satisfy both time-window constraints imposed by customers and parking spaces.

We initially reduce the redundant length of $U_i$ for each parking space $i$. For instance, if a parking space $i$ is available for an MPL from 10:00 to 16:00 ($[E_i, L_i]$=[10:00,16:00]), but no customer time windows within the range between 14:00 and 16:00 ($[E_{nk}, L_{nk}] = [14:00,16:00]$), $U_i$ can be reduced since $[E_i, L_i]$ is now only lasts from 10:00 to 14:00. The available time interval of parking space $i$ can be reduced to $U_i'$:

$$U_i' = \{[E_i, L_i] \cap [E_{nk}, L_{nk}] \quad \forall n \in N; \ k \in \{1, 2, \cdots, \|K_n\|\}; \ i \in I\} \tag{15}$$

To further reduce the range of available time intervals, we determine the earliest time window and the latest time window among all customers within each parking space. The optimal range of available time intervals will be updated as $U_i^{opt}$. By defining $U_i^{opt}$, we obtain a set of sub-intervals (Equation 17) of parking space $i$ when executing $U_i^{opt}$ divided by the service time $SV_i$:

$$U_i^{opt} = \{[min\{E_{1k}, \cdots, E_{nk}\}, max\{L_{1k}, \cdots, L_{nk}\}] \cap U_i' \quad \forall n \in N; k \in \{1, \cdots, \|K_n\|\}\} \tag{16}$$

$$[e_{ia}, l_{ia}] = U_i^{opt} / SV_i \qquad \forall i \in I; a \in \{1, 2, \cdots, \|U_i\|\}; \ [e_{ia}, l_{ia}] \in U_i^{opt} \tag{17}$$

Thus, for a single task $O_{ima}$ fulfilled by MPL $m$ at parking space $i$ within the $a^{th}$ sub-interval can be represented by Equation 18, where $q_{ia}$ denotes the total demand at the $a^{th}$ sub-interval at parking space $i$. Following this logic, we can obtain a task list $O$ that includes all subtasks $O_{ima}$ once we enumerate all parking spaces.

$$O_{ima} = \{(q_{ia}, m, i, [e_{ia}, l_{ia}]) \quad \forall i \in I; \ m \in M; \ a \in \{1, 2, \cdots, \|U_i\|\} \} \tag{18}$$

### 4.2 The structure of a state

By obtaining the tasks list for all MPLs, we can now construct the state of HQM (Figure 6). Each column in the variable $x_1$ and $x_2$ corresponds to a subtask $O_{ima} \in O$, i.e., the length of either $x_1$ or $x_2$ denotes the total number of tasks $\|O\|$ to be performed by MPLs. Each element of variable $x_1$ represents the unique identification number (ID) of an MPL $m$ that perform the corresponding subtask. Once the variable $x_1$ is determined, we can introduce $\bar{O}_m$ to represent the task list that assigns to a specific MPL $m$ ($m \in M$) where $O_{ima} \in \bar{O}_m$ is satisfied. The elements within the variable $x_2$ denote the unique task ID for subtask $O_{ima}$ so that forms the task execution sequence. The element of the variable $x_1$ is produced by a random integer generator with the range from 0 to $\|M\|$, while the elements of the variable $x_2$ is generated within the range from 0 to $\|O\|$. It is permissible to assign the same MPL to different subtasks $O_{ima}$, therefore the value of elements can be the same within the variable $x_1$. However, the task ID should never be repeated, as each subtask $O_{ima}$ is distinct and is executed only once by an MPL $m$.

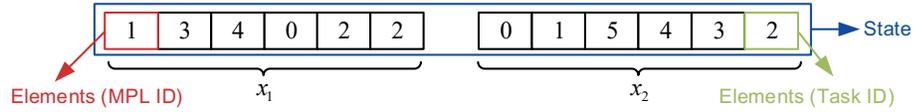

**Fig. 6** Illustration of a state

### 4.3 Hybrid Q-Learning network model (HQM)

We introduce an HQM that combines global and local search to generate a set of feasible solutions and obtains the optimal result by updating the Q-value iteratively.

In addition to the state $s = (x_1, x_2)$ defined previously, we define $\pi = (s_1, \cdots, s_p)$ as an agent that consists of multiple states $s$, which provides a set of feasible solution for the problem. The action $A_t$ is defined as a decision behaviour in which the agent $\pi$ selects a policy to obtain a better solution/state $s$ in the next timestep. $A_t$ is sampled from the right-hand side of Equation (20) during training time and obtained by $\varepsilon-$ **greedy** search of the Q-Learning and global-local search mechanisms. Our goal is to maximise the reward $R$ (Equation 19) obtained from a set of generated MPL route $G$:

$$R(\pi|G) = \max \left\{ \frac{1}{f_{s_1}(x_m, y_{ijm}, z_{ima})}, \cdots, \frac{1}{f_{s_p}(x_m, y_{ijm}, z_{ima})} \right\} \quad s_p \in \|\pi\| \tag{19}$$

The proposed HQM is parameterised by $\theta_{x_1}$ and $\theta_{x_2}$, corresponding to $x_1$ and $x_2$ respectively. It updates a strategy $p(\pi|G)$ that promotes agent $\pi$ deriving the optimal solution/state $s^*$. It is a stochastic policy and can be factorised as:

$$p_{\theta_{x_1}, \theta_{x_2}}(\pi|G) = \prod_{i=1}^{t} p_{\theta_{x_1}, \theta_{x_2}}[\pi(t)|\pi(<t), G] \tag{20}$$

To maximise the objective in Equation (19), we update every component on the right-hand side of Equation (20) based on the Q-value matrix in every timestep $t$. A new/updated state $s'$ in the next timestep $t+1$ will resemble the current state $s$ with the maximum Q-value in the current timestep $t$, indicating that the new state $s'$ retains the elements that contributed significantly to the optimal solution in the current timestep $t$. To accelerate the search for the optimal solution/state $s^*$, we generate multiple agents $\pi$ for the global search.

### 4.3.1 Encoder

We develop an encoder that consists of a random generator and a permutation pool to produce the agent $\pi$ as a normalised the input of HQM. As shown in the elements in Figure 7, each element within $x_1$ and $x_2$ is represented by $\theta_x^o$ ($x \in \{x_1, x_2\}$, $o \in \{1, \cdots, O\}$). The value next to the elements of $x_1$ and $x_2$ represents the probability of each current element that is remain selected by the current subtask $o$ in the next timestep $t+1$, and will be updated by the attention layer embedded within the decoder at each timestep $t$ (will introduce in *Attention-based decoder* Section).

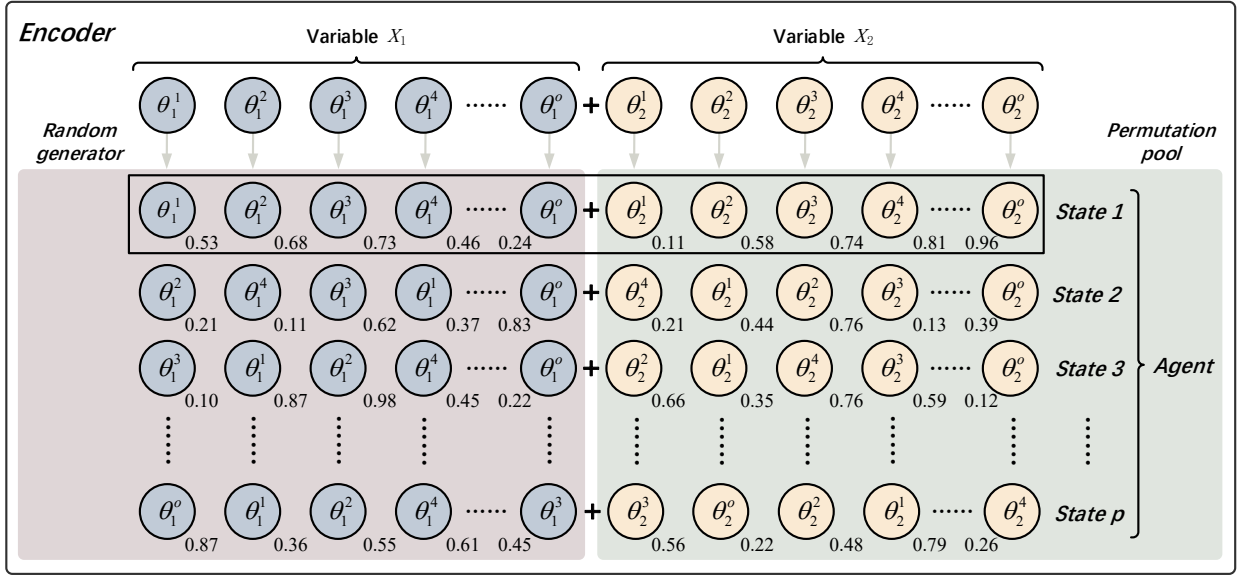

**Fig. 7** Structure of the encoder

### 4.3.2 Route generation & adjustment strategies

The route generation mechanism is based on utilising agent $\pi$ produced by the encoder. Specifically, each state $s$ within the agent $\pi$ corresponds to the route for MPLs. However, since the state $s$ is generated randomly, the resulting route may violate time-window requirements of tasks. We then introduce route adjustment strategies to resolve such time-window conflicts that occur during the route generation phase.

#### Mechanism of route generation

The route generation assigns and sequences tasks for MPLs based on the principle of subtask-MPL paring within the state $s$ (as shown in Figure 6). The resulting route will be applied in the global & local search mechanism (Section 4.3.4) for adjustment such that obtain better solution (as shown in the global & local search diagram in Figure 5). To capture the route generation process for MPLs, we modified the lemma developed by Schwerdfeger and Boysen (2020) and prove that for any given subtask $O_{ima}$, the task starting/ending time can be adjusted to execute the task as early as possible such that forms an optimal task $O_{ima}^*$.

**Lemma 1.** *For any subtask $O_{ima}$ defined in (18), there is an optimal task starting time $A^*_{ima}$ and ending time $\hat{A}^*_{ima}$ adjusted from $O_{ima}$ to form an optimal task $O^*_{ima}$, where $A^*_{ima}$ and $\hat{A}^*_{ima}$ are as follows:*

*i) task starting time $A^*_{ima} = max\{\hat{A}_{(i-1)ma} + t_{(i-1)i}, e_{ia}\}$ $\quad \forall m \in M; a \in \{1, \cdots, \|U_i\|\}, e_{ia} \in [e_{ia}, l_{ia}];$*

*ii) task ending time $\hat{A}^*_{ima} = \max\{\max\{A_{ima}, E_{nk}\} + SV_i\} \forall m \in M; n \in N; a \in \{1, \cdots, \|U_i\|\}; E_{nk} \in [E_{nk}, L_{nk}].$*

**Proof:** To simplify the process, we assume that each customer $n$ only has a single location ($K_n = 1$), but our proof is still valid for the multiple-locations situation. Following the nomenclature in Section 3.2, the optimal starting time of MPL $m$ in the first parking space is:

$$A_{1ma} = \max\{e_{1a}, t_{01}\} \qquad \forall m \in M; a \in \{1, \cdots, \|U_i\|\}; e_{1a} \in [e_{1a}, l_{1a}] \tag{21}$$

Since the MPL $m$ departs from the depot to the first parking space, the optimal starting time of MPL $m$ is the maximum value between the travel time and the earliest available task starting time in the first parking space.

We assume that the MPL $m$ serves the customers $n$ as early as possible. Thus, MPL $m$ can leave the first parking space as soon as the task is completed:

$$\hat{A}_{1ma} = \{\max(A_{1ma}, E_{nk}) + SV_1\} \quad \forall m \in M; n \in N; a \in \{1, \cdots, \|U_i\|\}; E_{nk} \in [E_{nk}, L_{nk}] \tag{22}$$

Similarly, the earliest task starting time and ending time in the second parking space are:

$$A_{2ma} = \max\{\hat{A}_{1ma} + t_{12}, e_{2a}\} \qquad \forall m \in M; a \in \{1, \cdots, \|U_i\|\}, e_{2a} \in [e_{2a}, l_{2a}] \tag{23}$$

$$\hat{A}_{2ma} = \{\max(A_{2ma}, E_{nk}) + SV_2\} \quad \forall m \in M; n \in N; a \in \{1, \cdots, \|U_i\|\}; E_{nk} \in [E_{nk}, L_{nk}] \tag{24}$$

Following the above logic, the earliest task starting $A^*_{ima}$ and ending time $\hat{A}^*_{ima}$ for each subtask $O_{ima}$ can be represented as:

$$A^*_{ima} = \max\{\hat{A}_{(i-1)ma} + t_{(i-1)i}, e_{ia}\} \qquad \forall m \in M; a \in \{1, \cdots, \|U_i\|\}, e_{ia} \in [e_{ia}, l_{ia}] \tag{25}$$

$$\hat{A}^*_{ima} = \max\{\max\{A_{ima}, E_{nk}\} + SV_i\} \quad \forall m \in M; n \in N; a \in \{1, \cdots, \|U_i\|\}; E_{nk} \in [E_{nk}, L_{nk}] \tag{26}$$

*Route adjustment strategy and route generation algorithm*

The route generator described in the previous section does not consider the capacity constraint and stochastic events (e.g., traffic congestion, service delay) in real implementation, which violates the constraint outlined in Equation (9) and the time-window requirement of a single task, respectively. Thus, a route adjustment strategy is necessary to ensure the feasibility of all MPL routes. Algorithm 1 outlines the procedure of route generations and adjustments.

The route adjustment strategy consists of two algorithms: a back-to-depot (BTD) strategy, and a holding-at-current-parking-space (HCPS) option. The former allows the MPL to reload parcel and postpone deliveries to meet the time-window and capacity constraints. However, this strategy results in an extra distance, impacting the operating costs and delivery efficiency. This process is described in Equation (27) where $t_{i0}$ and $t_{0j}$ represent the time consumption of MPL from parking space $i$ to parking space $j$ via the central depot.

The latter enables MPLs to wait at their current parking space, postponing the start time of the next task. Thus, the MPL $m$ will remain at the current parking space $i$ until the latest ending time $l_{ia}$ of the $a$th sub-interval, as shown in Equation (28). In contrast to BTD, HCPS is only used to resolve time-window conflicts.

$$A_{jma} = A_{ima} + SV_i + t_{i0} + t_{0j} \quad \forall i, j \in I; m \in M; a \in \{1, 2, \cdots, \|U_i\|\} \tag{27}$$

$$A_{jma} = l_{ia} + t_{ij} \quad \forall i, j \in I; m \in M; a \in \{1, 2, \cdots, \|U_i\|\} \tag{28}$$

---

**Algorithm 1:** Route generation and adjustment algorithm

---

Input: Task set $\bar{O}_m$ for $m \in M$; Locker speed $v$; Locker capacity $Q$

output: The delivery route of MPLs

1  Initialise $A_{ima}, q_{ia}, e_{ia}, l_{ia}, E_i, L_i, SV_i$

2  Calculate travel time $t_{ij}$ between parking space $i$ and $j$ for all parking spaces

3  **for** each subtask $O_{ima}$ *in* $\bar{O}_m$ **do**

4      Add depot to current task list $\bar{O}_m$ // Depart from depot

5      *visited_route_list* $\leftarrow \emptyset$ // Record the route of MPLs

6      **for** *next in* $O_{ima}$ **do**

7          *current_parking_space* $\leftarrow \bar{O}_m[O_{ima}]$ //Map the current task of MPL $m$ to parking space

8          *next_parking_space* $\leftarrow \bar{O}_m[next]$ //Map the next task of MPL $m$ to the next parking space

9          Add *current_parking_space* to *visited_route_list*

10         Update $A_{ima}, q_{ia}, e_{ia}, l_{ia}$

11         *travel_time* $\leftarrow$ lialg.norm(*current_parking_space* $-$ *next_parking_space*) / $v$

12         **if** $A_{ima} + SV_i + t_{ij} < e_{ja}$ *and* $l_{ja} + t_{ij} \geq e_{ja}$ **then** // Violate time-window constraint of tasks

13             Adjust task starting and ending time based on ***Equation (28)*** // Apply HCPS

14             Update $A_{jma}, q_{ja}, e_{ja}, l_{ja}, E_j, L_j, SV_j$

15             Add *next_parking_space* to *visited_route_list*

16         **else if** $A_{jma} < e_{ja}$ *or* $q_{ia} > Q$ **then** // Violate time windows or constraint (9)

17             Adjust task starting and ending time based on ***Equation (27)*** // Apply BTD

18             Update $A_{jma}, q_{ja}, e_{ja}, l_{ja}, E_j, L_j, SV_j$

19             Add *next_parking_space* to *visited_route_list*

20         **else if** *satisfy constraints* **then**

21             Add *next_parking_space* to *visited_route_list*

22         **end if**

23     **end for**

24 **end for**

25 **return** *visited_route_list*

---

*4.3.3 Attention-based decoder*

The attention-based decoder is developed to calculate the reward $R$ of the route generated in *Mechanism of route generation* Section and updated the probability of each current element that is remain selected by the current subtask $O_{ima}$ in the next timestep $t + 1$ through the Q-value matrix embedded within the attention layer (see *Mechanism of attention layer* Section). The reward calculation is based on Equation (19), given the route generated.

Figure 8a to 8d demonstrate the decoding procedure for an agent $\pi$. As shown in Figure 8a, the decoder initially decomposes the first state $s_1$ in the timestep $t + 1$ by applying a decoding algorithm (see Appendix A) and calculating the corresponding reward $R$, followed by updating the probability of each element within the state $s_1$. Analogously, the decoding procedure lasts until all states within the agent $\pi$ have been completed (Figure 8b and 8c). Finally, all states within the decoder are sorted according to the reward obtained, and the optimal state $s^*$ is then extracted to determine the optimal delivery route by executing Algorithm 1 (Figure 8d).

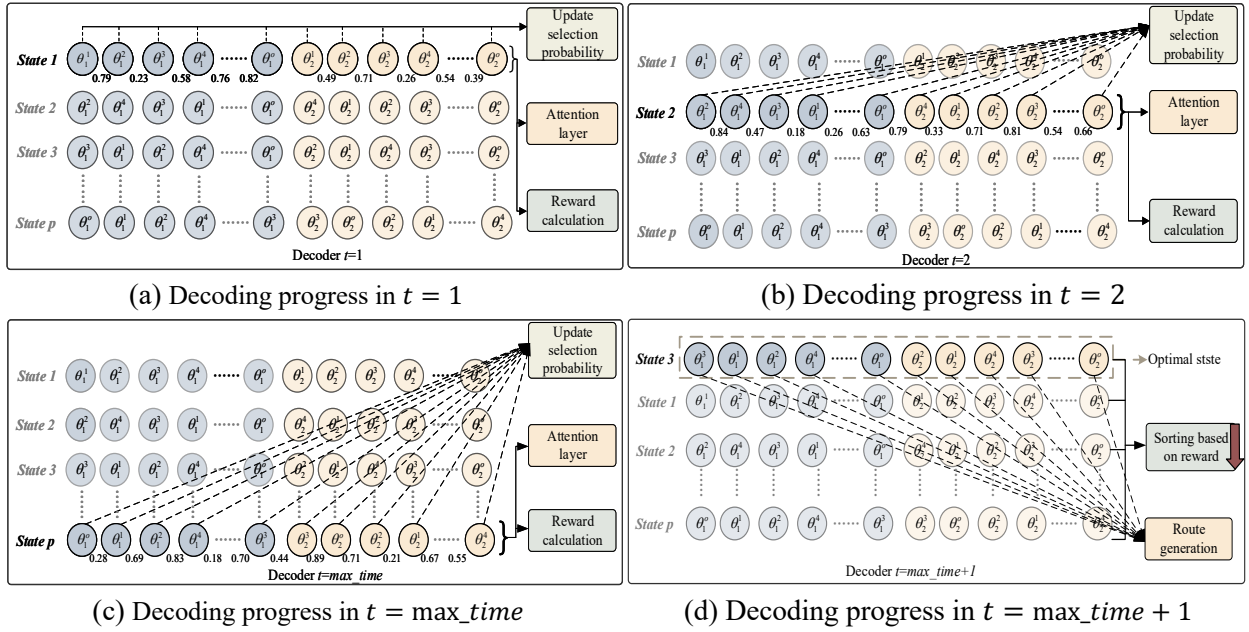

(a) Decoding progress in $t = 1$  (b) Decoding progress in $t = 2$

(c) Decoding progress in $t = \text{max\_time}$  (d) Decoding progress in $t = \text{max\_time} + 1$

**Fig. 8** Structure of the decoder

*Mechanism of attention layer*

The state-action space of HQM can be represented by a lookup table (Figure 9). The size of the lookup table equals $O \times T$, where $O$ equals the length of the task list of a state $s$, while $T$ equals the total number of stepsize/episode of the whole HQM process. Given this property, the size of the lookup table increases with the problem size exponentially.

The attention mechanism is developed to tackle such a curse of dimensionality, in which we decompose the state-action space into multiple low-dimensional state-action combination chains linked by the Q-value matrix. We define two Q-value matrices in Equations (29) and (30), which correspond to $x_1$ and $x_2$ within a state $s$, respectively. Each value within the matrix reflects the contribution of the element to the final reward.

Furthermore, each value within the matrix indicates the interdependence between two adjacent actions, where the larger the value, the closer the two adjacent actions are connected. The new state $s'$ updated in the next action $A_{t+1}$ inherits the elite elements (the elements that contribute to the greater reward) of the current state $s$, which have greater Q-values in the current action $A_t$. Therefore, the solution can be enhanced while inheriting the elite elements of the current state $s$.

$$Q\left(\theta_{x_1}^o\right) = \begin{bmatrix} Q_{11}\left(\theta_{x_1}^1\right) & \cdots & Q_{1m}\left(\theta_{x_1}^o\right) \\ \vdots & \ddots & \vdots \\ Q_{m1}\left(\theta_{x_1}^1\right) & \cdots & Q_{mm}\left(\theta_{x_1}^o\right) \end{bmatrix} \quad \forall o \in \{1, \cdots, O\}, m \in M \tag{29}$$

$$Q\left(\theta_{x_2}^o\right) = \begin{bmatrix} Q_{11}\left(\theta_{x_2}^1\right) & \cdots & Q_{1o}\left(\theta_{x_2}^o\right) \\ \vdots & \ddots & \vdots \\ Q_{o1}\left(\theta_{x_2}^1\right) & \cdots & Q_{oo}\left(\theta_{x_2}^o\right) \end{bmatrix} \quad \forall o \in \{1, \cdots, O\} \tag{30}$$

Based on the Q-value matrices defined above, we can determine the possibility of the current element being selected for the current task $o$ in the following timestep $t + 1$, denote as $p(\theta_{x_1}^o)$ and $p(\theta_{x_2}^o)$. These are calculated as follows:

$$p(\theta_{x_1}^o) = \frac{exp\,(Q(\theta_{x_1}^o))}{\sum_{o=1}^O exp\,(Q(\theta_{x_1}^o))} \qquad o \in \{1, \cdots, O\} \tag{31}$$

$$p(\theta_{x_2}^o) = \frac{exp\,(Q(\theta_{x_2}^o))}{\sum_{i=1}^O exp\,(Q(\theta_{x_2}^o))} \qquad o \in \{1, \cdots, O\} \tag{32}$$

*Mechanism of Q-value update*

To update the Q-value matrix in every timestep $t$, we propose the following Q-value function for each element of a state $s$ within an agent $\pi$:

$$Q_{t+1}^{\theta_{x_i}}(s_t^{x_i}, a_t^{x_i}) = Q_t^{\theta_{x_i}}(s_t^{x_i}, a_t^{x_i}) + \alpha \left[ R_t(s_t^{x_i}, s_{t+1}^{x_i}, a_t^{x_i}) + \gamma \max_{a^{x_i} \in A_t} Q_t^{\theta_{x_i}}(s_{t+1}^{x_i}, a^{x_i}) - Q_t^{\theta_{x_i}}(s_t^{x_i}, a_t^{x_i}) \right] \tag{33}$$

where $x_i \in \{x_1, x_2\}$ and $o \in O$ , $R_t(s_t^{x_i}, s_{t+1}^{x_i}, a_t^{x_i})$ represents the reward obtained from state $s_t^{x_i}$ by executing action $a_t^{x_i}$ towards a state $s_{t+1}^{x_i}$ in timestep $t$, $\alpha$ represents the learning rate, and $\gamma$ represents the discount factor.

The update procedure can be regarded as a Markov decision process that derives satisfactory results through updating the Q-value between two adjacent actions. For each action $a_t^{x_i} \in A_t$ ($t \in \{1, \cdots, max\_time\}, x_i \in \{x_1, x_2\}$) of the variable, there is a corresponding Q-value ($Q_t^{\theta_{x_1}}$ and $Q_t^{\theta_{x_2}}$) associated with action $a_t^{x_i}$. We use $A_t$ to denote the action profile that integrates both actions of variable $x_1$ ($a_t^{x_1}$) and $x_2$ ($a_t^{x_2}$) in timestep $t$ (i.e., $A_t = a_t^{x_1} \cup a_t^{x_2}$).

The action $A_t$ selects the direction of migration that will seek a better solution in the next timestep $t + 1$ based on a Q-value such that migrate into the new policy. Once $A_t$ is determined and executed by the agent $\pi$, and the next action $A_{t+1}$ will be selected according to the estimated $Q_{t+1}$ (Figure 9).

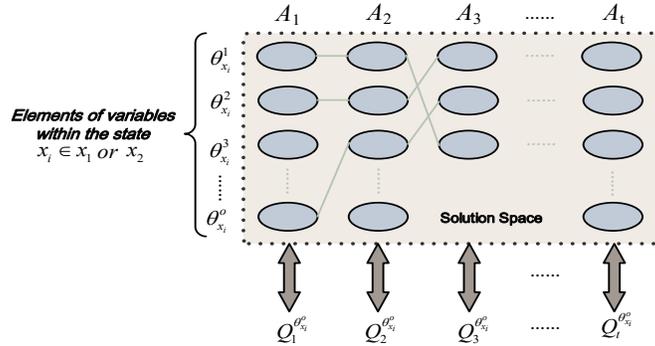

**Fig. 9** Lookup table for state-action space

*4.3.4 Policy for action selection*

Q-Learning suffers from the trade-off between exploitation and exploration of the information feedback obtained from previous actions (Sutton and Barto, 2018). A strong emphasis on information exploration may derive the optimal solution but may sacrifice the speed of convergence. While favouring information exploitation can improve convergence speed at the expense of optimality.

As part of our approach, we propose integrating the Q-value-based global search with a local search approach to defining the policy for action selection, leading to a near-optimal solution while improving convergence speed. Concretely, the global search mechanism first updates each element within the state $s$ of the agent $\pi$ to generate a new state $s'$ given a specific action $A_t$, in which the resulting reward is denoted as $R'$. Next, the local search mechanism adjusts the subset of elements within an individual state $s'$ updated by the global mechanism to create another new state $s''$, in which the resulting reward is represented as $R''$. The HQM outputs either $s'$ or $s''$ for the final solution follows $max\{R', R''\}$.

By implementing a combined global and local search approach, the agent $\pi$ and corresponding Q-value matrices will be updated accordingly. The iteration stops once the error of the Q-value matrices generated by two adjacent actions $A_t$ and $A_{t+1}$ is less than $1 \times 10^{-8}$. The state $s^*$ with the greatest reward $R^*$ among all agent $\pi$ is defined as the optimal solution of MPLP.

In terms of global search mechanism, the migration between the two adjacent actions $A_{t-1}, A_t$ updates the agent $\pi$ based on the Q-value obtained from the previous action $A_{t-1}$. The policy of action selection is based on the $\varepsilon - greedy$ search which can be defined as follows:

$$a_{t+1}^{x_i} = \begin{cases} \underset{a_t^{x_i} \in a_t}{\text{argmax}}\, Q_t^{\theta_{x_i}^0}\left(s_{t+1}^{x_i}, a_t^{x_i}\right), & \tau < \varepsilon \\ a_t^{x_i{'}}, & \tau \geq \varepsilon \end{cases} \tag{34}$$

Where $\varepsilon$ denotes the greedy factor, $\tau$ denotes the random value between 0 and 1, and $a_t^{x_i{'}}$ denotes the action that is being selected based on the normalised Q-value matrices within the global scope. The greater the Q-value is, the greater probability of the corresponding element inheriting an elite fragment of the current state $s$.

The local search operation focuses on adjusting the subset of the variable $x_1$ and $x_2$ within an individual state $s_i$ to create a new state $s_i'$. The adjustment for the elements within variable $x_1$ is represented by the following:

$$s_i'(x_1) = \left| s_i(x_1) + \vec{\omega} \cdot \left( s_i(x_1) - s_{neighbour}(x_1) \right) \right| \tag{35}$$

where $s_{neighbour}$ represent the neighbour state in $\{s_{i+1}, s_{i-1}\}$. $\vec{\omega}$ is a random vector, where each element $o$ and $o'$ within the two corresponding neighbour state $s_{neighbor}$ is assigned to a value based on the following piecewise function. This allows the modification of only subsets of the elements in $x_1$. Since all elements within a state $s$ should be non-negative, we take absolute values for all elements within the new state $s_i'(x_1)$.

$$\vec{\omega} = \begin{cases} -1 \text{ or } 1 & \text{if } o < o' \\ 0 & \text{otherwise} \end{cases} \tag{36}$$

Since the elements within the variable $x_2$ is unique and non-repeated, the adjustments for the elements within variable $x_2$ focus on the subset exchange within the same state to guarantee the uniqueness of elements (Figure 10). The start points and end points are generated randomly follows $X_2 \sim U(0, \|O\|)$. We obtain a new state $s_i'(x_1, x_2)$ by integrating new variable $s_i'(x_1)$ and $s_i'(x_2)$.

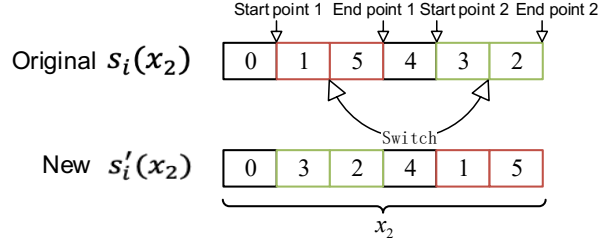

**Fig. 10** Illustration of adjustment for variable $x_2$

The implementation procedure of HQM is shown in Algorithm 2.

---

**Algorithm 2:** Hybrid Q-Learning network method (HQM)

---

Input: Number of MPLs $\|M\|$; All task list $O$; Number of agents $Agent$; Timestep $T$

output: The optimal state $s = (x_1, x_2)$

1 *Initialise learning rate $\alpha$, discount factor $\gamma$, random value $\tau, \omega$, greedy factor $\epsilon$, Q-value matrices $Q_t^{\theta_1^Q}, Q_t^{\theta_2^Q}$*

2 **for** $i$ ***in*** range($Agent$) **do** // Initialise agents

3      $x_1 \leftarrow \mathbb{N} \sim U(0, \|M\|)$

4      $x_2 \leftarrow$ Permutating task list $O$

5      $s = (x_1, x_2)$ // Generate initial state $s$

6      Execute ***Decoding algorithm, Algorithm 1***

7 **end for**

8 Rank and record all state $s$ based on the reward obtained

9 **for** $t$ ***in*** range($T$) **do**

10      Record the current best state $s^*$ and the corresponding reward $R^*$ of the agent $\pi$

11      **for** $o$ ***in*** range($O$) **do**

12          Normalise Q-value matrices $Q_t^{\theta_1^Q}, Q_t^{\theta_2^Q}$

13      **end for**

14      **for** $i$ ***in*** range($Agent$) **do** // Global search

15          **for** $o$ ***in*** range($O$) **do**

16              $s'(x_1', x_2') \leftarrow$ ***Equation (35)*** // Implement global search and obtain new state $s'$

17              Execute ***Decoding algorithm, Algorithm 1***

18          **end for**

19          $R' \leftarrow$ ***Equation (19)*** // Calculate the reward $R'$ for the new state $s'$

20          **if** $R' > R^*$ **then**

21              $s^*, R^* \leftarrow s', R'$ // Update the optimal state $s^*$ and reward $R^*$

22          **end if**

23      **end for**

24      **for** $i$ ***in*** range($Agent$) **do** // Local search

25          $s''(x_1'', x_2'') \leftarrow$ ***Equation (36)*** // Implement local search and obtain another new state $s''$

26          Execute ***Decoding algorithm, Algorithm 1***

27          $R'' \leftarrow$ ***Equation (19)*** // Calculate the reward $R''$ for the new state $s''$

28          **if** $R'' > R^*$ **then**

29              $s^*, R^* \leftarrow s'', R''$ // Update the optimal state $s^*$ and reward $R^*$

30          **end if**

---

```
31 │  end for
32 │  $Q_t^{\theta_1^Q}, Q_t^{\theta_2^Q} \leftarrow$ **Equation (33)** // Update Q-value matrices
33 │  if *satisfy convergence constraints* then
34 │  │  break
35 │  end if
36 end for
37 return $s^*$
```

## 5 Results and discussion

We evaluate the performance of HQM for the proposed MPLP. To validate the proposed mathematical model and identify the solution gap between HQM and the exact approach counterpart, we first compare HQM's derived solution and computation time with the exact approach. We then apply a Wilcoxon signed-rank to evaluate the result significance between exact approach and HQM. Since the main purpose of our study is to emphasise the advantage of HQM in solving large problem instances compared with meta-heuristic approaches, we next compared the proposed HQM with GA based on the following criteria: a) the performance indicators derived from the objective function (Equation 4), b) the demand fulfilment rate, c) the performance of the solution improvement, d) reward acquisition, e) learning convergence speed and f) the computing efficiency. The HQM vs. GA experiments are all executed under both BTD & HCP strategies are applied.

Beyond the performance comparison between HQM and GA, we further investigate the effect of key factors on influencing different cost units of the objective function. Lastly, we provide managerial insights and fruitful discussions regarding the implementation of our method to benefit academia and the logistics profession.

### 5.1 Generation of problem instance and parameter setting

Benchmark instances commonly used for VRP experiments (e.g., Solomon problem instances) are not suitable for our MPLP, since the latter involves customers' trajectory within a day, which is out of the scope of the VRP. On the other hand, since prior MPLP studies (Li et al., 2021; Schwerdfeger and Boysen, 2020) neither consider the dynamic route adjustment technique as ours nor disclose the data set of problem instances and precise optimal solution of different cost units, we cannot compare the derived results obtained by HQM with the existing studies under the same dimension. Therefore, we create our own problem instance and analyse the impact of network size on the deployment and operation of MPLs. We provide the problem instance, detailed solutions, and relevant code of experiment to ensure the rigour of our research and contribute to future MPLP & VRP studies (see https://github.com/Yubin-Liu/Hybrid-Q-Learning-Network-Approach-for-MPLP.git).

In total, 96 problem instances are evaluated based on 24 networks. Each network has a predefined number of MPL parking locations (ranging from $I_i = 5$ to $I_i = 10$), and a set of customer locations $N_n$ close to these $I_i$ ($N_n = \{5, 10, 15, 20\}$).

The HQM and GA code are implemented in Python 3.8.5 on an Apple M1 (3.20 GHz) processor with 16GB RAM, while the exact approach is implemented on the same machine via Gurobi solver. To ensure that the results are comparable, we configured the two algorithms' implementation under the same timestep/iteration and agent/population size, and with similar parameters in terms of learning rate/elite rate and mutation rate to the degree possible. The corresponding HQM and GA parameters are presented in Table 4. The setup of the problem instance is as follows:

- **Generation of customer demands:** The number of parcels per customer $n \in N$ is defined as a uniform random distribution $U \sim (1, 4)$.

- **Generation of locations:** For each location of customer $n \in N$, the coordinates are randomly generated within a specified MPL radius $r_m$ of a parking space $i \in I_i$ to easily form different clusters, in other words $d_{ni}^k \leq r_m$ given that $d_{ni}^k = \sqrt{(x_{nk} - x_i)^2 + (y_{nk} - y_i)^2}$ ($x_i \sim U(0, x_{max})$, $y_i \sim U(0, y_{max})$), where $d_{ni}^k$ denotes the distance between the parking space $i$ and the $k$th stopover of the customer $n$.

- **Customer stopover generation:** The number of stopovers $\|K_n\|$ of a customer $n$ within the planning horizontal follows $\|K_n\| \sim U(1, 3)$. The maximum walking distance $\rho_n$ (km/customer) of customer $n$ is generated by following normal distribution $\rho_n \sim N(0.5, 0.1)$ (Schwerdfeger and Boysen, 2020).

- **Generation of time interval:** We generate non-overlapping time intervals for each customer $n$. The time interval $[E_{nk}, L_{nk}]$ of the $k$th locations of a customer $n$ is generated by a random generator, in which $E_{nk}$, $L_{nk} \in T_{list} = \{8 \times 60, \cdots, 18 \times 60\}$ (corresponding to working hours from 8:00 to 18:00) with the length of $T_{list}$ equals 10. Meanwhile, we guarantee that the interval length of time window $|L_{nk} - E_{nk}|$ follows normal distribution $|L_{nk} - E_{nk}| \sim N(60, 5)$, subject to $L_{nk} > E_{nk}$. The generation of parking spaces' time intervals is similar to the above procedure with an interval length $|L_i - E_i|$ following normal distribution $|L_i - E_i| \sim N(50, 5)$. We ensure that at least one parking space suits each customer.

**Table 4.** Parameters setting in model validation

| Parameter | Symbol | Value |
|---|---|---|
| **Parameters in mathematical model** | | |
| Service radius of MPLs | $r_m[km]$ | 5 |
| Capacity of MPLs | $Q[unit]$ | 20 |
| Travel speed of MPLs | $v[km/hr]$ | 40 |
| Fixed cost of locker $m$ | $F_m[\pounds/locker]$ | 20,000 |
| Unit cost travelling from parking space $i$ to $j$ | $C_{ij}[\pounds/km]$ | 0.5 |
| A large indefinite number | $\phi[-]$ | $10^8$ |
| Service time of parking space $i$ | $SV_i[min]$ | 10 |
| Weights of cost units | $W_1 : W_2 : W_3 [-]$ | 10:1:5 |
| **Parameters in customer properties** | | |
| Maximum walking range | $\rho_n[km]$ | $\rho_n \sim N$ (0.5, 0.1) |
| Length of time interval of customer | $T_s = |L_{nk} - E_{nk}|[min]$ | $T_s \sim N$ (60, 5) |
| **Parameters in parking spaces** | | |
| Length of time interval of parking spaces | $T_p = |E_i, L_i|[min]$ | $T_p \sim N$ (50, 5) |
| Service buffer time | $\delta[min]$ | 10 |
| **Parameters in HQM** | | |
| Number of actions/timestep | $A_t[-]$ | 1000 |
| Learning rate | $\alpha[-]$ | $0.9 \times e^{-\left(\frac{current\ timestep}{total\ timestep}\right)}$ |
| Discount factor | $\gamma[-]$ | 0.9 |
| Greedy factor | $\varepsilon[-]$ | $U(0,1)$ |
| Random variables | $\tau, \omega[-]$ | $U(0,1)$ |
| Total states within an agent | $Agent[-]$ | 100 |
| **Parameters in GA** | | |
| Number of iteration/timestep | $Iteration[-]$ | 1000 |
| Population size | $Pop[-]$ | 100 |
| Possibility of crossover | $P_{cross}[\%/Pop]$ | 50 |
| Proportion of elite size | $P_{elite}[\%/Pop]$ | 5 |
| Mutation rate | $P_{muta}[\%/Pop]$ | 5 |

**Notes:** MPLs' fleet size and service delay are given higher weightings because the deployment of MPLs requires high initial costs, while the service delay has a significant effect on the market share for parcel deliveries (Iwan, Kijewska and Lemke, 2016). Nevertheless, the weights of different cost units may be adjusted in accordance with real operations.

*5.2 Performance comparison between HQM and exact approach*

We utilise the Gurobi optimiser to solve the problem instances via the exact approach, and the obtained results are compared with the HQM. The solver was terminated after 2 hours or when it has found an optimal solution. The comparison is evaluated based on two aspects: i) the best value of the objective function, cost units, and the computation time; ii) the significance between the results obtained by Gurobi solver and HQM.

Table 5 shows the best reward obtained by Gurobi solver and HQM. Although the Gurobi solver generates superior solutions for small problem sizes (e.g., nodes ≤ 100), the computation time is relatively long (average computation time = 3199.74s). While HQM achieve high-quality solutions in large problem instances with significant shorter computation time (average computation time = 609.05s). The average best result between HQM and Gurobi is close with an average gap of 6.21%. Such a gap is influenced by the extreme values caused by HQM obtaining lower rewards in small problem instances (since the solving accuracy of HQM requires larger sample). In real operations, however, we conclude that HQM can achieve better results in shorter computation time than the Gurobi solver based on the results shown in Table 5.

Appendix C shows the best value of different cost units of Gurobi solver and HQM. We notice that the Gurobi solver achieves less MPLs deployment in small problem instances while HQM obtains greater distance savings in large problem instances. Specifically, the difference between the Gurobi solver and HQM in MPLs deployment is within ±3 units. Regarding travel distance, HQM achieves a 9.93% improvement compared with that of the Gurobi solver in general. There is a minor difference between the Gurobi solver and HQM on average service delays (9.07 min of HQM vs. 10.49 min of Gurobi).

**Table 5.** Comparison of average computation time under different problem instances

| Number of parking spaces $I_i$ | Customer locations within each parking space $N_n$ | Gurobi | | HQM | | Gap (%) |
|---|---|---|---|---|---|---|
| | | Computation Time (Sec) | Best Result ($\times 10^{-3}$) | Computation Time (Sec) | Average Final Reward ($\times 10^{-3}$) | |
| 5 | 5 | 896.5 | 7.247 | 264.3 | 6.481 | -10.57 |
| | 10 | 1232.4 | 4.204 | 384.6 | 3.899 | -7.26 |
| | 15 | 1381.9 | 3.930 | 433.5 | 3.655 | -6.99 |
| | 20 | 3497.2 | 1.096 | 506.4 | 1.158 | 5.66 |
| 6 | 5 | 898.6 | 5.774 | 309 | 5.149 | -10.82 |
| | 10 | 1371.5 | 3.078 | 438.9 | 2.697 | -12.38 |
| | 15 | 2491.8 | 2.162 | 499.5 | 2.024 | -6.38 |
| | 20 | 3671.6 | 2.063 | 580.5 | 2.227 | 7.95 |
| 7 | 5 | 677.1 | 4.672 | 367.2 | 4.281 | -8.37 |
| | 10 | 1625.6 | 0.695 | 551.7 | 0.687 | -1.15 |
| | 15 | 2884.6 | 0.383 | 701.4 | 0.427 | 11.49 |
| | 20 | 4965.9 | 0.408 | 757.8 | 0.456 | 11.76 |
| 8 | 5 | 948.2 | 4.062 | 384.3 | 3.828 | -5.76 |
| | 10 | 2995.3 | 0.675 | 562.5 | 0.717 | 6.22 |
| | 15 | 4543.7 | 0.317 | 734.7 | 0.354 | 11.67 |
| | 20 | 5779.9 | 0.252 | 951.9 | 0.264 | 4.76 |
| 9 | 5 | 971.5 | 3.772 | 411.9 | 3.478 | -7.79 |
| | 10 | 3126.5 | 0.357 | 751.2 | 0.325 | -8.96 |
| | 15 | 6079.1 | 0.639 | 856.8 | 0.709 | 10.96 |
| | 20 | 7200 | 0.251 | 978.9 | 0.274 | 9.16 |
| 10 | 5 | 1324.3 | 2.728 | 444.6 | 2.539 | -6.93 |
| | 10 | 4796.8 | 0.339 | 701.4 | 0.360 | 6.19 |

| | 15 | 6233.7 | 0.240 | 914.4 | 0.275 | 14.58 |
| | 20 | 7200 | 0.130 | 1129.8 | 0.139 | 6.92 |
| Mean | | 3199.74 | 2.061 | 609.05 | 1.933 | -6.21 |

**Notes:** i) *Average Final Reward* = [$R(BTD) + R(HCPS)$]/2; ii) To ensure that the results are comparable, we transpose ***Best Result*** = 1 / *Equation (3)*; iii) ***Gap*** = (***Average Final Reward*** of HQM - ***Best Result*** of Gurobi) / ***Best Result*** of Gurobi

We conducted statistical tests to evaluate the performance of HQM statistically. Wilcoxon signed-rank tests were used to determine the significance of the difference between two groups of results obtained by HQM and the Gurobi solver. Wilcoxon signed-rank is a non-parametric statistical test to analyse two related samples (Walpole et al., 2017) and was used for a VRP problem (Yu et al., 2022b). Tables 6 shows the result for Wilcoxon signed-rank tests in terms of the best reward and computation time, respectively. We used the confidence level of alpha = 0.05 to test the hypothesis. It is concluded that the two paired samples differ significantly if the significance level ($p - value$) is less than alpha.

In Table 6, the $p - value$ of the best result comparison is larger than alpha, implying that the performance of the HQM is close to the Gurobi solver in terms of solution quality. Additionally, we conclude that the computation time of HQM and the Gurobi solver are significantly different since the $p - value$ is less than alpha. Finally, as a result of Tables 5 and 6, we infer that the HQM significantly outperforms the the Gurobi solver in terms of solution quality and computation time for large problem instances.

**Table 6.** Result of Wilcoxon signed-rank test on best result obtained and computation time

| Indicator <br> Sample | Test on Best Result/Reward Obtained | | Test on Computation Time | |
|---|---|---|---|---|
| | $Z$ | $p - value$ (2-tailed) | $Z$ | $p - value$ (2-tailed) |
| HQM vs. Gurobi | 1.5 | 0.145 | -4.3 | 0.000* |

**Notes:** * denotes the significant difference exists.

### 5.3 Performance comparison between HQM and GA

We next compared the performance of HQM with GA regarding the optimal configuration of MPLs (fleet size, travel distance, and service delay) and optimisation performance (reward acquisition, convergence speed, and computation time). We also apply advanced techniques (e.g., roulette wheel selection, elite strategy) to different operations within GA to ensure good computation performance and rigour in comparisons (the implement procedure of GA see Appendix B).

### 5.3.1 Optimal configuration of MPLs

We first investigate the implementation results in fleet size and travel distance under different problem instances (see Appendix D). The evaluation is based on two dimensions: i) the performance of algorithms, and ii) the performance of route adjustment strategies. Our results indicate that the maximum difference in achieving optimal fleet size between HQM and GA is $\pm 3$ units (the maximum MPL fleet size is 34 units). In fact, the GA achieves a better performance than the HQM in terms of minimising driving distances but causes higher service delay significantly.

Regarding the performance of BTD and HCPS route adjustment strategies, BTD achieves fewer MPL deployment in small-median size problem instances on average ($I_i \leq 8$, $N_n \leq 15$). However, such fleet size difference is compromised in large networks ($I_i \geq 9$) because HCPS is regarded as a more cost-efficient approach to resolve time window conflicts compared with deploying dense MPLs caused by BTD. With the increase in parking spaces and customer locations, the HCPS strategy allows for greater driving distance savings than BTD since fewer round trips are required between parking spaces and the central depot.

In terms of average service delay, the results obtained from the two algorithms are 10.10min (HQM-BTD), 8.91min (HQM-HCPS), 121.03min (GA-BTD), and 113.77min (GA-HCPS), respectively. The average service delay obtained by HQM is therefore significantly lower than the GA's, contributing to higher reward (Figure 11). GA, however, shows randomness and non-robust performance in optimising service delay across different network sizes, as GA produces high latency even at small problem instances (Figure 11c and 11d), while HQM enables on-time delivery in small network sizes (Figure 11a and 11b). Comparatively to BTD implemented by HQM and GA, HCPS reduces delays by 11.85% (HQM) and 6.00% (BTD), respectively.

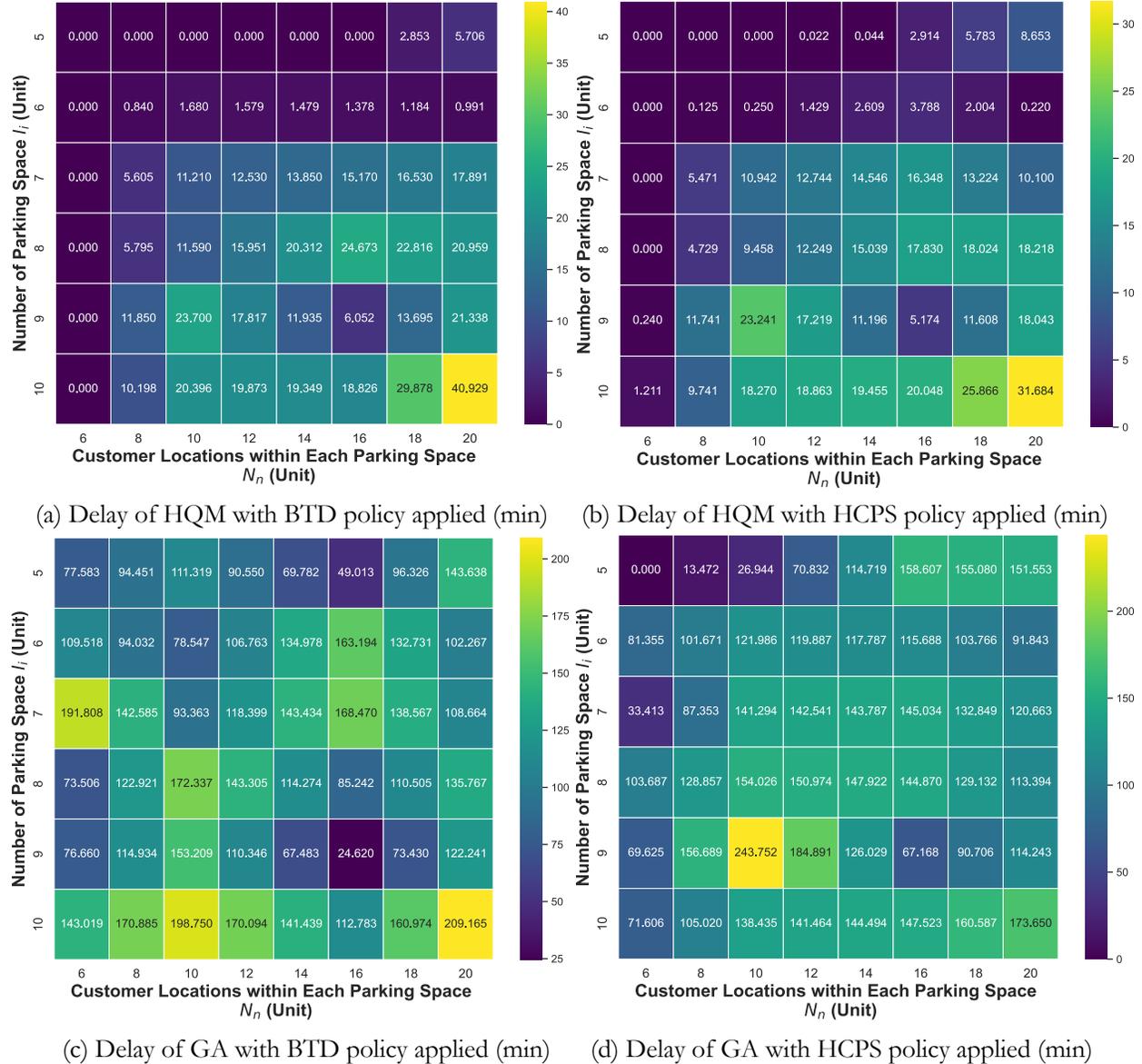

(a) Delay of HQM with BTD policy applied (min)  (b) Delay of HQM with HCPS policy applied (min)

(c) Delay of GA with BTD policy applied (min)  (d) Delay of GA with HCPS policy applied (min)

**Fig. 11** Service delay of HQM and GA with different route adjustment polices applied

We further evaluate the demand fulfilment in the first-delivery round, i.e., the maximum demand that an MPL can fulfil in a single delivery trip. Operators anticipate a higher first-time delivery rate with smaller fleet sizes and shorter travel distances, leading to better utilisation of delivery fleets and reducing repeated deliveries.

Table 7 shows that HQM performs better than the GA, as HQM reduces the return to depot frequency. Concretely, the average first-time delivery fulfilment of HQM ranges from 15.5 units to 19.25 units (the maximum capacity is 20 units per MPL), corresponding to the fulfilment rate from 77.50% to 96.30% respectively. In addition, HCPS strategy obtains higher first-time demand fulfilment that BTD when $I_i \geq 8$.

**Table 7.** Parcel first-time fulfilment under HQM and GA implementation

| Number of parking spaces $I_i$ | Locations within each parking space $N_n$ | HQM | | GA | |
|---|---|---|---|---|---|
| | | Parcel first-time fulfilment-BTD (units) | Parcel first-time fulfilment-HCPS (units) | Parcel first-time fulfilment-BTD (units) | Parcel first-time fulfilment-HCPS (units) |
| 5 | 5 | 13 | 7 | 7 | 6 |
| | 10 | 11 | 15 | 9 | 15 |
| | 15 | 18 | 20 | 18 | 18 |
| | 20 | 18 | 20 | 17 | 17 |
| | Mean value | 15 | 15.5 | 12.75 | 14 |
| 6 | 5 | 11 | 12 | 10 | 9 |
| | 10 | 17 | 19 | 13 | 14 |
| | 15 | 19 | 20 | 18 | 18 |
| | 20 | 20 | 16 | 19 | 19 |
| | Mean value | 16.75 | 16.75 | 15 | 15 |
| 7 | 5 | 11 | 10 | 9 | 9 |
| | 10 | 18 | 15 | 14 | 16 |
| | 15 | 19 | 14 | 19 | 20 |
| | 20 | 18 | 20 | 18 | 18 |
| | Mean value | 16.5 | 14.75 | 15 | 15.75 |
| 8 | 5 | 16 | 17 | 10 | 12 |
| | 10 | 14 | 17 | 18 | 18 |
| | 15 | 20 | 20 | 19 | 18 |
| | 20 | 19 | 20 | 17 | 20 |
| | Mean value | 14.75 | 18.5 | 16 | 17 |
| 9 | 5 | 10 | 11 | 9 | 9 |
| | 10 | 18 | 20 | 16 | 18 |
| | 15 | 20 | 20 | 20 | 15 |
| | 20 | 20 | 20 | 20 | 20 |
| | Mean value | 17 | 17.75 | 16.25 | 15.5 |
| 10 | 5 | 20 | 14 | 14 | 15 |
| | 10 | 20 | 20 | 15 | 17 |
| | 15 | 19 | 20 | 19 | 16 |
| | 20 | 18 | 20 | 20 | 20 |
| | Mean value | 19.25 | 18.5 | 17 | 17 |

*5.3.2 Optimisation performance of HQM*

The optimisation performance of HQM is evaluated from three dimensions: (i) the final reward obtained, (ii) reward acquisition and convergence, and (iii) computation time. The final reward obtained represents the absolute ability to derive a final optimal solution, while the reward acquisition and convergence speed measure the improvement rate throughout the learning process.

*Final reward obtained*

Figure 12 shows the final reward obtained from HQM and GA under different problem instances and route adjustment strategies. We conclude that the proposed HQM combined with HCPS strategy yields the greatest reward with the value of $1.97 \times 10^{-3}$, followed by $1.90 \times 10^{-3}$ (HQM-BTD), $1 \times 10^{-3}$ (GA-HCPS), and $0.97 \times 10^{-3}$ (GA-BTD).

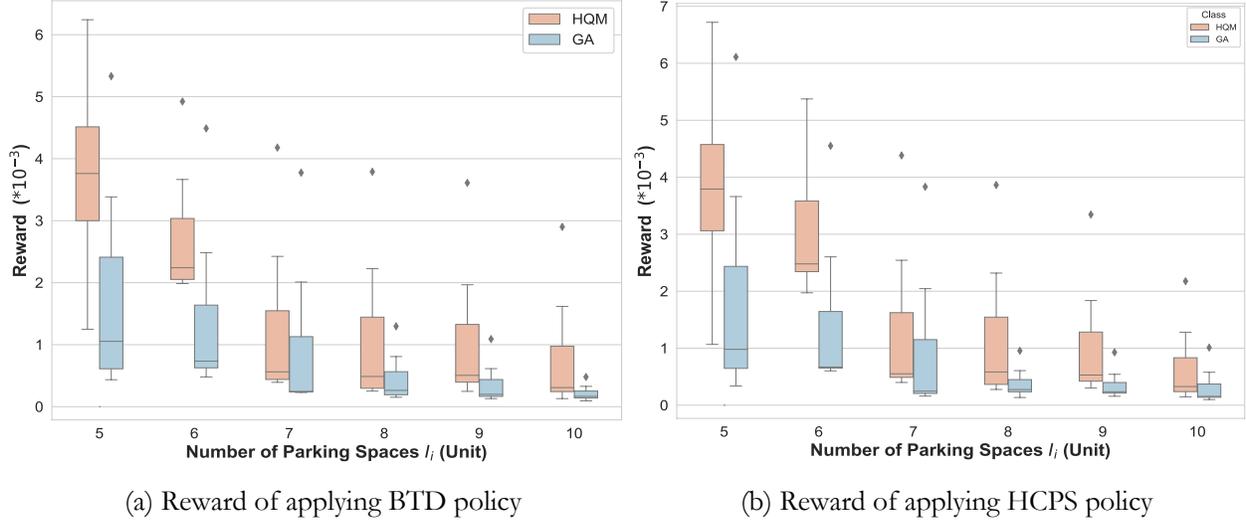

(a) Reward of applying BTD policy      (b) Reward of applying HCPS policy

**Fig. 12** Reward obtained under different problem instances

The surface plot in Figure 13 illustrates the reward improvement of HQM and GA, where the bottom layer denotes the initial rewards, and the top layer denotes the final rewards optimised by the two algorithms. The calculation of the improvement based on the metric: $(R_{t=\max} - R_{t=1})/R_{t=1}$. In Figure 13b, the larger gap between the two layers within the small and medium problem instances ($I_i \leq 8, N_n \leq 18$) indicates that the HQM-HCPS achieves a greater reward improvement than the HQM-BTD (Figure 13a). Contrary to the HQM, the GA is merely good at solving small problem instances ($I_i \leq 7, N_n \leq 10$) and performed poorly when medium and large problem instances are involved (Figure 13c and 13d). Although HQM shows less reward improvement in large problem instances compared with small-medium networks due to the large gap of axis calibration in the figure, the HQM still perform better than GA, which can be proven by the concrete improvement rate shown in Figure 14.

Specifically, HQM-HCPS achieves an average increase of 453.58% in final rewards over initial rewards (Figure 14b), while HQM-BTD achieves an average increase of 433.23% (Figure 14a). By contrast, GA-HCPS achieves an average improvement of 97.31% in final rewards compared to its initial rewards (Figure 14d), while GA-BTD achieves an average improvement of 92.50% (Figure 14c). The optimisation ability of HQM is on average 348.49% greater than that of GA, suggesting that HQM facilitates an efficient search for better solutions and overcomes local optima.

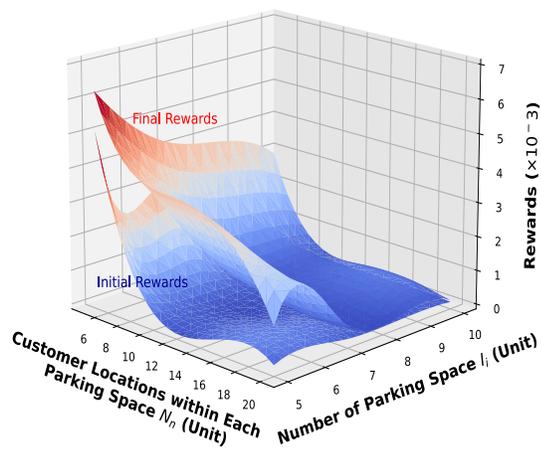

(a) Reward comparison of HQM-BTD

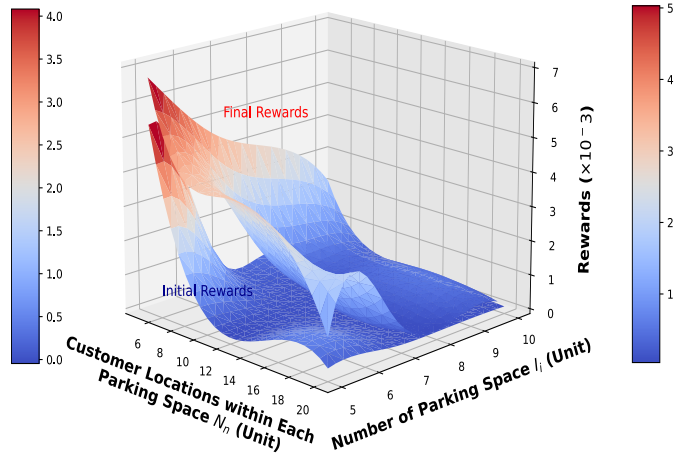

(b) Reward comparison of HQM-HCPS

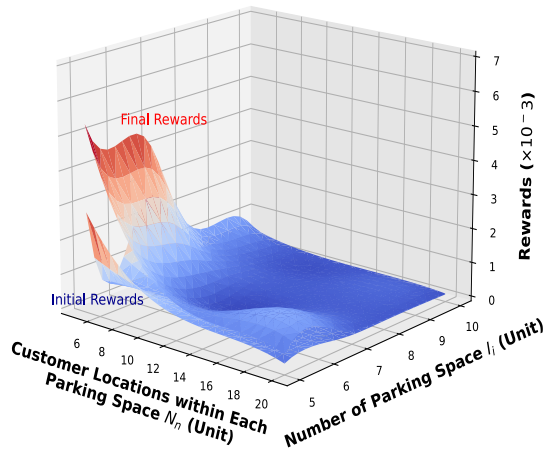

(c) Reward comparison of GA-BTD

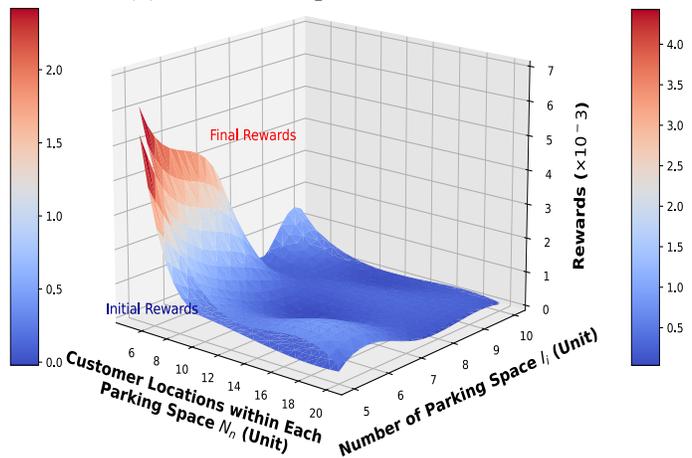

(d) Reward comparison of GA-HCPS

**Fig. 13** Optimisation ability of HQM and GA

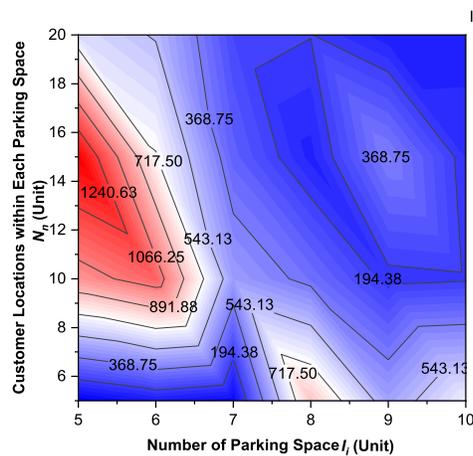

(a) Solution improvement of HQM-BTD

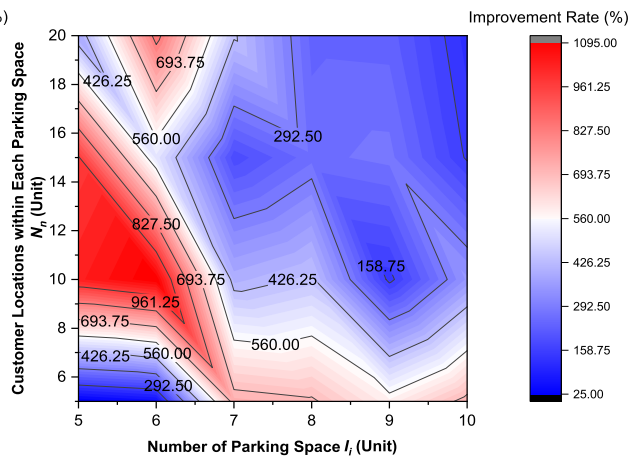

(b) Solution improvement of HQM-HCPS

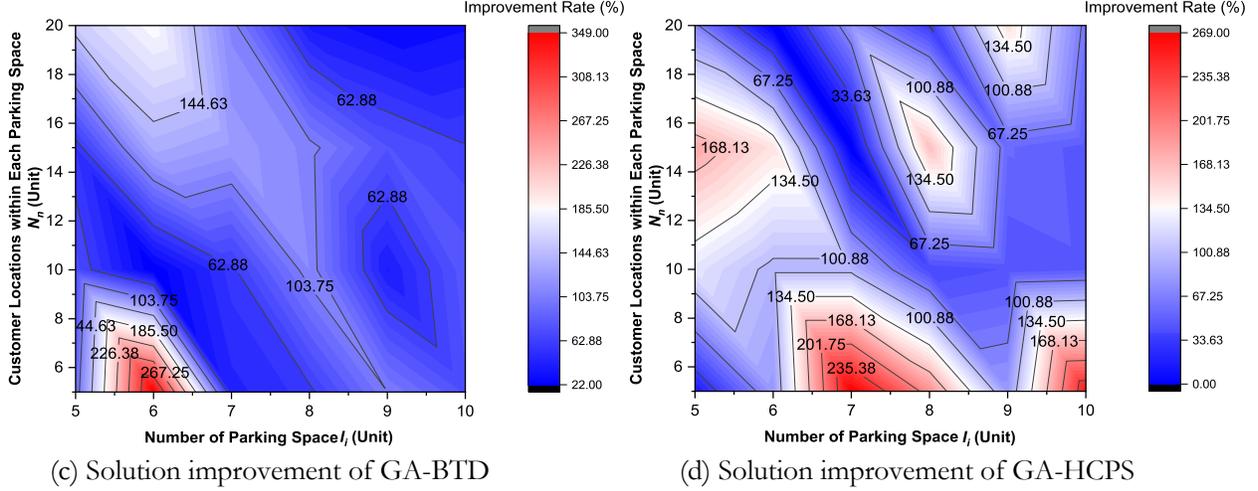

(c) Solution improvement of GA-BTD      (d) Solution improvement of GA-HCPS

**Fig. 14** Improvement rate of HQM and GA

*Reward acquisition and convergence*

Figure 15 illustrates the reward acquisition and convergence of HQM and GA. Algorithms are implemented under a large instance setting ($I_i = 10$, $N_n = 20$). The result concludes that GA tends to fall into a local optimum during the earlier iterations, resulting in a lower reward (Figure 15a). The reason for this can be explained by Figure 15b, which illustrates the convergence performance of the algorithms: we can see that the objective value of GA-BTD and GA-HCPS sharply converges at generation 60 (±20) and generation 200 (±10) respectively, whereas HQM improves solution smoothly and converges at generation 180 (±20), meaning that HQM better exploits the information feedback obtained from the previous action and generate more efficient solution search policy than GA.

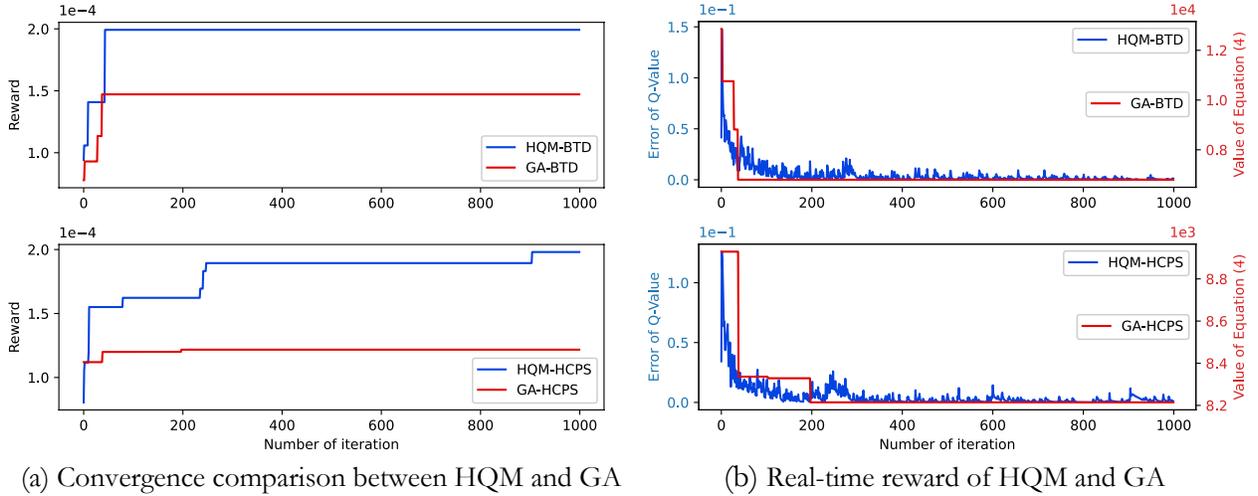

(a) Convergence comparison between HQM and GA      (b) Real-time reward of HQM and GA

**Fig. 15** Reward acquisition and convergence comparison

*Computing time under different problem instances*

In Table 8, we compare the computation time between HQM and GA. The results indicate that the average computation time of HQM is 9.64 times that of GA given the same problem instance. This is due to the fact that the heuristic algorithm does not involve data sampling and knowledge mining, thus reducing the computation time required, but sacrifices satisfactory solutions (the final reward of HQM is greater than GA's for all problem instances).

Since there are no MPLP benchmark problem instances introduced and solved by RL-based approaches, we cannot compare HQM's computation time with other RL-based approaches within the same dimension. However, in contrast to other attention-mechanism-based RL approaches for solving VRP with similar network sizes, our proposed HQM provides a shorter computation cycle. We here cite the related research conducted by Kool et al. (2018) for reference which has been proven that the proposed REINFORCE approach achieves better optimisation performance than other RL techniques: the author proposed a model based on attention layers with benefits over the Pointer Network. The REINFORCE is applied to train the model. The computation time under different problem instances with varied sizes is as follows: i) CVRP: 6 min (Node=20), 28 min (Node=50), 2 hr (Node =100); ii) SDVRP: 9 min (Node =20), 42 min (Node =50), 3 hr (Node =100).

**Table 8.** Comparison of average computation time under different problem instances

| Number of parking spaces $I_i$ | Customer locations within each parking space $N_n$ | HQM | | GA | | Gap (%) |
|---|---|---|---|---|---|---|
| | | Computation Time (Sec) | Average Final Reward ($\times 10^{-3}$) | Computation Time (Sec) | Average Final Reward ($\times 10^{-3}$) | |
| 5 | 5 | 264.3 | 6.481 | 23.7 | 5.721 | 11.73 |
| | 10 | 384.6 | 3.899 | 48.3 | 1.325 | 66.02 |
| | 15 | 433.5 | 3.655 | 26.7 | 0.710 | 80.57 |
| | 20 | 506.4 | 1.158 | 30 | 0.386 | 66.67 |
| 6 | 5 | 309 | 5.149 | 42 | 4.520 | 12.22 |
| | 10 | 438.9 | 2.697 | 46.2 | 0.568 | 78.94 |
| | 15 | 499.5 | 2.024 | 57.9 | 0.681 | 66.35 |
| | 20 | 580.5 | 2.227 | 48 | 0.698 | 68.66 |
| 7 | 5 | 367.2 | 4.281 | 23.4 | 3.804 | 11.14 |
| | 10 | 551.7 | 0.687 | 46.8 | 0.252 | 63.32 |
| | 15 | 701.4 | 0.427 | 44.7 | 0.195 | 54.33 |
| | 20 | 757.8 | 0.456 | 85.8 | 0.247 | 45.83 |
| 8 | 5 | 384.3 | 3.828 | 32.4 | 1.127 | 70.56 |
| | 10 | 562.5 | 0.717 | 48 | 0.289 | 59.69 |
| | 15 | 734.7 | 0.354 | 61.5 | 0.249 | 29.66 |
| | 20 | 951.9 | 0.264 | 109.8 | 0.144 | 45.45 |
| 9 | 5 | 411.9 | 3.478 | 29.7 | 1.010 | 70.96 |
| | 10 | 751.2 | 0.325 | 60.9 | 0.147 | 54.77 |
| | 15 | 856.8 | 0.709 | 126.3 | 0.258 | 63.61 |
| | 20 | 978.9 | 0.274 | 108 | 0.174 | 36.50 |
| 10 | 5 | 444.6 | 2.539 | 47.4 | 0.746 | 70.62 |
| | 10 | 701.4 | 0.360 | 48.6 | 0.161 | 55.28 |
| | 15 | 914.4 | 0.275 | 157.8 | 0.160 | 41.82 |
| | 20 | 1129.8 | 0.139 | 117 | 0.097 | 30.22 |

**Notes:** *Gap* = (*Average Final Reward* of HQM - *Average Final Reward* of GA) / *Average Final Reward* of HQM

*5.4 Effects of critical factors against different cost units*

We eventually investigate the effects of the critical factors on three cost units within the objective function (fleet size, travel distance, and service delay) to provide managerial and operational insight to logistics professionals. Factors are classified into four dimensions, depending on their attributes: i) time windows from customers and parking spaces, ii) locker properties, iii) network density, and iv) locker service radius and customer maximum walking range. Sections 5.4.1 to 5.4.3 focus on a parametric discussion of how these factors influence different cost units, while Section 5.4.4 provides actionable managerial implications in practice.

### 5.4.1 Effects of critical factors against fleet size

Figure 16a demonstrates the effect of customer time windows and the available parking time of parking space on fleet size. The length of customer residence time $T_s$ and available parking time $T_p$ are evaluated within the range from 40 to 70 minutes. The result indicates that both customer time windows and the available parking time are negatively correlated to the fleet size. Extending customer residential time provides the potential to reduce time window conflicts, thus resulting in fewer MPLs being required for conflict resolution.

In terms of the effect of MPL properties, we notice that MPL capacity $Q$ has a more significant influence on fleet size than speed $V_I$ once $V_I \geq 40\ km/hr$ (Figure 16b). However, the fleet size will increase sharply if $V_I \leq 35\ km/hr$ and $Q \leq 15\ units$ because more dense contours are within this area. This can be explained by the fact that lower locker speed may not satisfy the time window requirements of customers and parking spaces, which require assigning extra MPLs to resolve such cases. In addition, a larger MPL fleet is required to meet time window constraints for lower MPL capacities. Our result suggests that the MPL capacity $Q$ is expected to be greater than 15 units/MPL in our network settings to reduce MPL fleet size.

Figure 16c shows that both the number of parking spaces and customer locations have conspicuous impacts on service delays, proven by the positive correlation of the contour plot. Compared with the other three factors, the difference in fleet size with varied network density is the most significant (the range of fleet size from 9 to 34 units).

Finally, we conclude that customers walking range $\rho_c$ have negligible impact on MPLs fleet size once the walking range $\rho_c \geq 0.4 km$ (Figure 16d), and that MPL service radius has a greater significance on fleet size. In summary, time window constraints, MPL capacity, and network density are the main factors affecting MPLs fleet size.

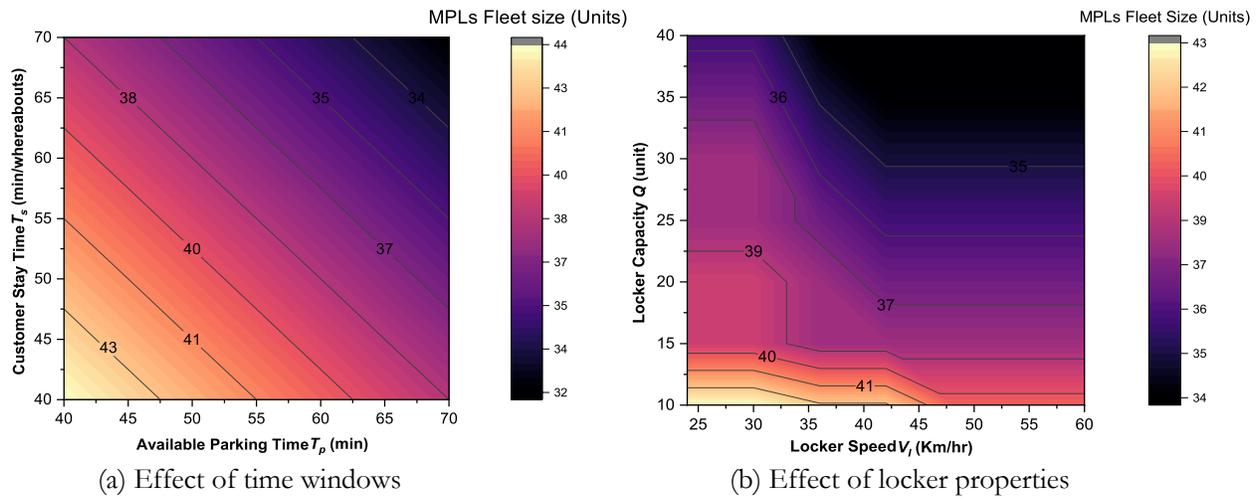

(a) Effect of time windows          (b) Effect of locker properties

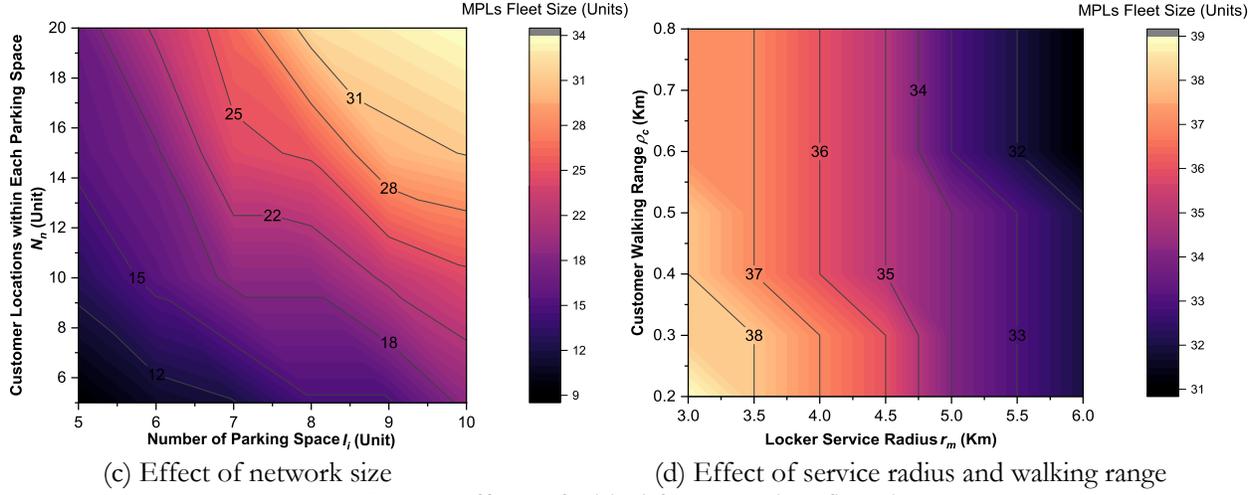

(c) Effect of network size      (d) Effect of service radius and walking range

**Fig. 16** Effects of critical factors against fleet size

*5.4.2 Effects of critical factors against travel distance*

The travel distance is more sensitive with available parking time $T_p$ compared with customer stay duration $T_s$ (Figure 17a). The travel distance sharply decreases when available parking time $T_p \geq 55min$, along with fewer fleet required to meet customer demands, thus leading to shorter routes. On the other hand, the travel distance has increased gradually when customer stay time $T_s \leq 60\ min$, because some MPLs are scheduled globally to meet the tight delivery time requirements even though the MPLs are far away from the customer to be fulfilled. This issue is even more significant when time windows constraints are tighter (e.g., $57min \leq T_s \leq 63min$, and $T_p \leq 45min$).

The result indicates that MPL capacity $Q$ is negatively correlated with travel distance, whereas MPL speed $V_l$ does not. The total route length only has a minor change when the locker capacity $Q > 20\ units/MPL$ (Figure 17b).

With the increased number of parking spaces and customer locations, the travel distance of MPLs has increased dramatically, especially the number of locations per parking space $N_n \geq 15$ units/parking space and $I_i \geq 8$ units (Figure 17c). Figure 17d demonstrates that the travel distance increased with their service range since more customers are covered by the MPL fleet that results in greater route length.

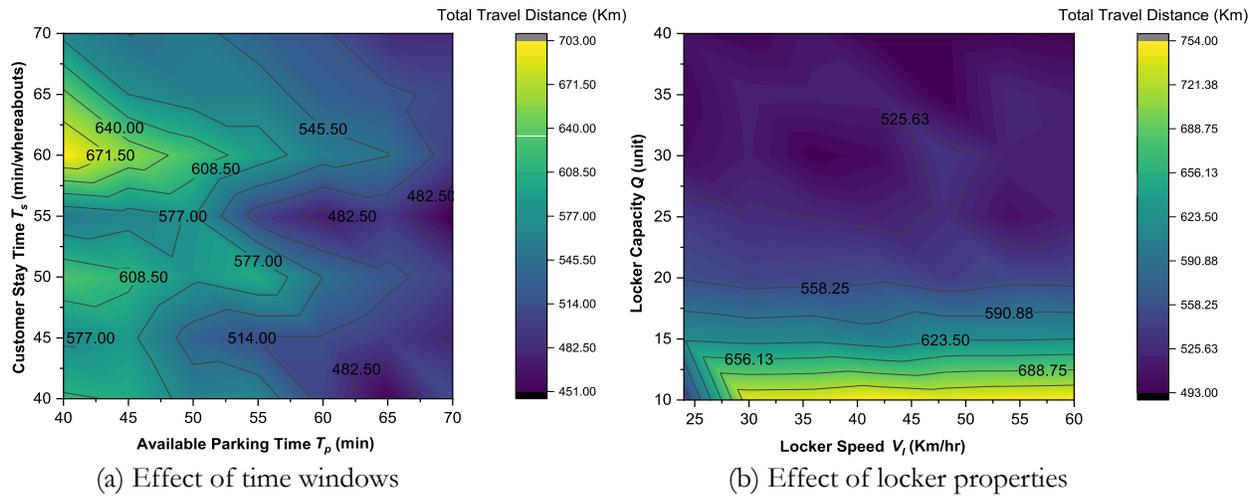

(a) Effect of time windows      (b) Effect of locker properties

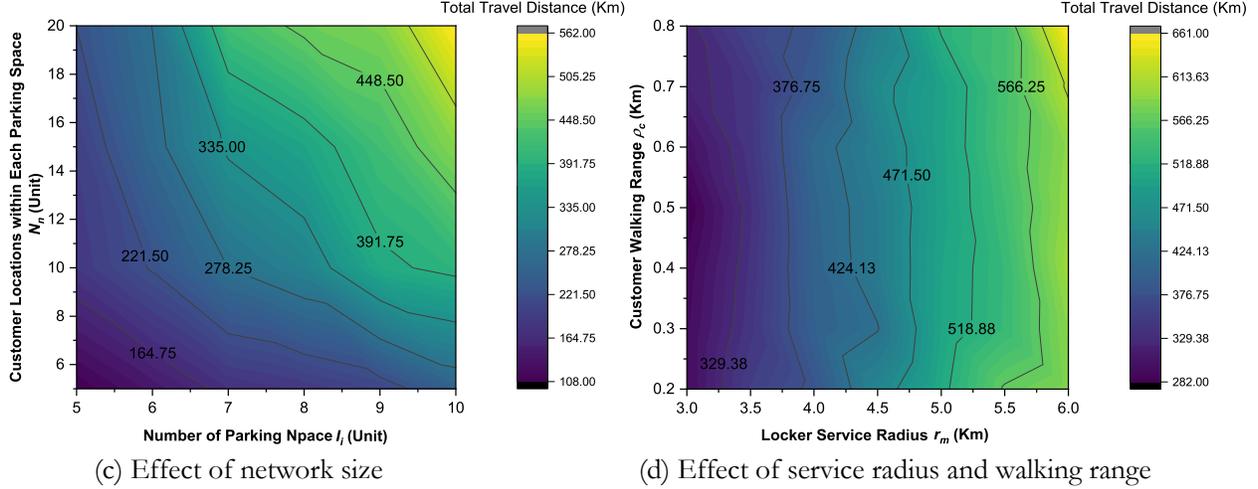

(c) Effect of network size      (d) Effect of service radius and walking range

**Fig. 17** Effects of critical factors against travel distance

*5.4.3 Effects of critical factors against service delay*

Figure 18a suggests that the service delay decreases with longer customer stay time $T_s$, meaning that if customers provide more generous time windows, the MPL can be scheduled more flexibly thus reducing service delays. Specifically, the service delay will be significantly reduced once $T_s \geq 50min$ or $T_p \geq 45min$. However, the service delay will exceed more than 60 min if both customers and parking spaces time windows are tight (e.g., $T_s < 50min$ and $T_p < 45min$). Furthermore, available parking time $T_p$ has a more significant impact on service delay compared with customer stay time $T_s$.

Figure 18b illustrates that the service delay will be improved as locker capacity $Q$ decreases. The reduced MPL capacity means fewer customers can be fulfilled in a single delivery round, which saves more travel distance and eliminates the potential delays. However, more MPLs are required to satisfy the remaining demands.

Conversely, the larger capacity $Q$ may promote the algorithm to assign more customers to the same MPL, thus reducing MPL fleet size with the expense of higher service delays. The service delay will be improved as MPL speed increases. However, if the MPL speed is too fast, the travel time between two adjacent parking spaces may be too short to satisfy the available parking time restriction in large problem instances. In this case, the MPL will adopt route adjustment strategies to resolve time window conflicts, which may result in higher delays in the whole system. We conclude that the capacity $Q \leq 25$ units may achieve satisfactory delay improvement since $Q = 25$ is a distinct boundary of the contour plot. This is of practical significance to logistics operators, as it guides operators to adopt MPLs with appropriate capacity to improve service delay in an economic fashion.

Figure 18c shows that the network sizes have positive conspicuous effects on service delay, especially in the case that $I_i \geq 9$ and $N_n \geq 16$. This is because the increase in parking spaces and customer locations leads to more MPL deployments and longer travel distances, resulting in higher delays. Finally, Figure 18d presents that the greater MPL service radius and customer walking distance will cause higher delays since more customers will be served within the service region. Given this concern, $r_m \leq 5km$ and $\rho_c \leq 0.6\ km$ achieve lower delays.

In summary, the time window constraints and network density are the crucial factors affecting service delay. This requires collaboration between logistics professionals, customers, and urban planners to identify the most appropriate settings for these parameters that achieve timely service at a reasonable cost. Additionally, carriers can improve customer satisfaction by defining a reasonable service radius and implementing effective scheduling policies based on customer preferences.

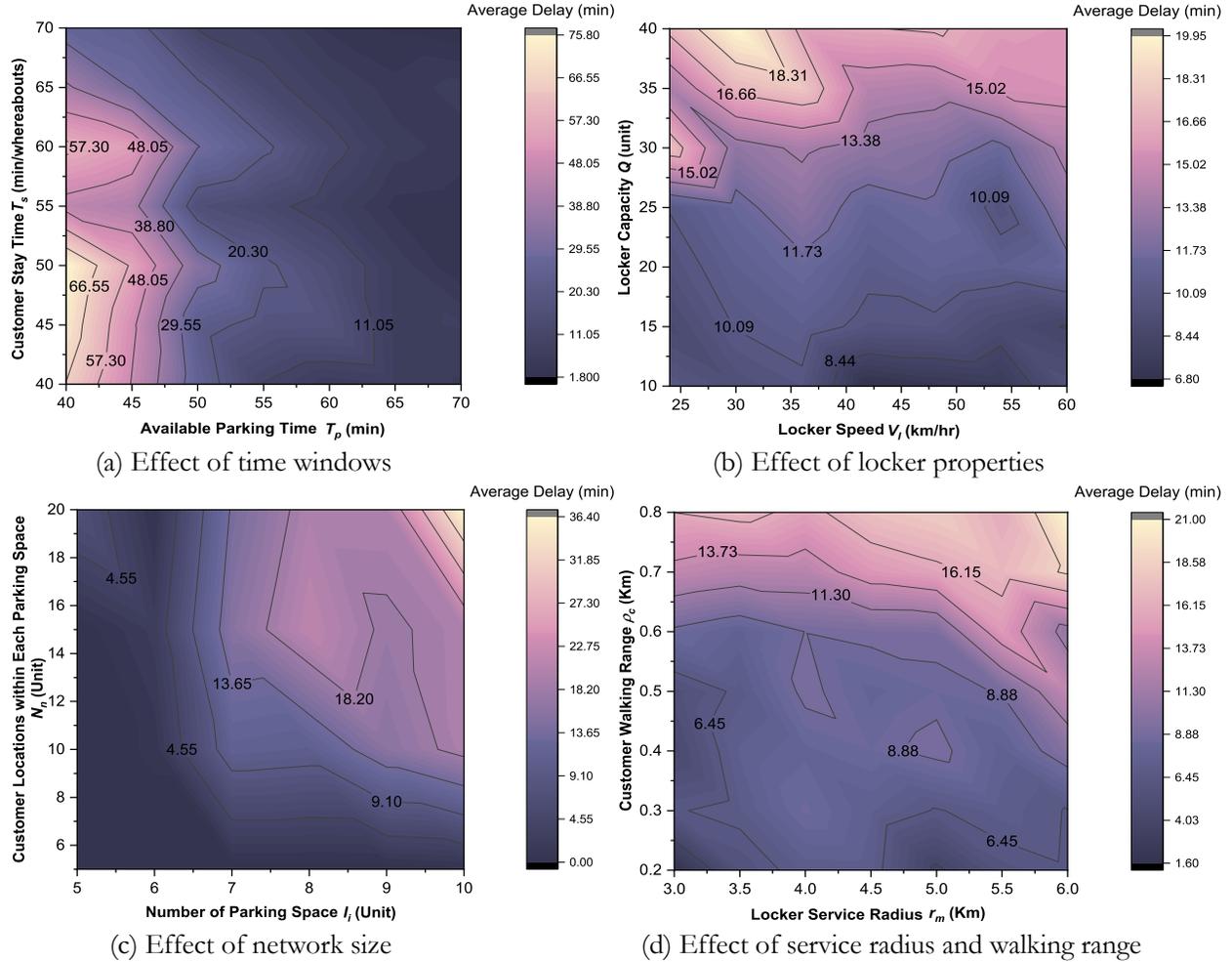

(a) Effect of time windows       (b) Effect of locker properties

(c) Effect of network size       (d) Effect of service radius and walking range

**Fig. 18** Effects of critical factors against service delay

*5.4.4 Managerial implications*

The managerial implications are discussed based on the prior parametric analysis (section 5.4.1 to 5.4.3) to enable logistics practitioners better deploying MPL operations at scale accordingly in the future.

*Customer time windows and available parking time*

Considering the tendency shown in Figures 16(a), Figures 17(a) and Figures 18(a), we conclude the ratio $\varpi$ of available parking time $T_p$ to customer stay time $T_s$ ($\varpi = T_p/T_s$) is regarded as a crucial indicator for logistics operators since the larger available parking time $T_p$ enable MPLs process more stopovers, thus reach more customers. For $\varpi \geq 1$, the performance of different cost units has significant improvement subject to $T_s, T_p \geq$ 60 min.

Nevertheless, from the perspective of customer stay time $T_s$, it is difficult for customers to offer a more generous time window in practice given that their schedules are often fixed during the weekdays, making it difficult for them to adapt time windows for parcel collection. Logistics operators will therefore require a more generous available parking time to fulfil orders more effectively. One possible solution would be to park MPLs in residential areas (there is less traffic on residential streets, so the MPLs can be parked for longer periods without causing significant congestion). As an alternative, the deployment of MPLs near work sites during peak commute hours would allow more customers to be served within a relatively concentrated time interval. Finally, the application of a track & trace system (e.g., smartphone app) allows customers to monitor MPL's position in real-time, thereby reducing parcel overlap time for parcel handover.

*Locker capacity and locker velocity*

Intuitively, greater MPL velocity contributes to more agility and can reach more customers by visiting more stops along the way. To achieve this, it would be possible to dispatch MPLs prior to the peak commute period, which would enable them to travel at a faster speed (e.g., $V_l \geq 35$km/hr in our case) in the curb-side lane, thereby reducing service delays due to the traffic congestion. However, it is important to note that this policy does not apply to most hours of the day, as there is also an increased demand during off-peak hours, particularly on weekends. Given this concern, increased locker speed on main roads or pavements is not a significant approach to improve delivery efficiency since the average vehicle speed in the urban areas of well-developed cities is limited to 25 km/hr (Kusumaningtyas and Lodewijks, 2008). Therefore, we recall the contour plot in Figure 16b, 16c, and 16d, the results suggest that the medium capacity level of MPLs (e.g., $Q \in [20, 30]$ units/MPL) achieves greater cost savings.

*Parking spaces and customer stopovers*

The location of the customer stopovers and demand density are important factors that influence the placement of parking spaces. Therefore, the successful adoption of the MPL concept depends on customer willingness to use MPL services and disclose their approximate location. Customers may have concerns about data privacy and the inconvenience of providing this information, which can be addressed by incentives offered by the logistics provider, such as reduced or waived service charges. Alternatively, logistics operators can offer different service options to customers who do not want to provide location information, such as stationary lockers or door-to-door delivery, while still providing MPL service to customers who are willing to use it.

From the perspective of demand density, it becomes easier to reach the maximum capacity of MPLs as demand density increases in an urban area. Hence, combining stationary lockers with MPL is a feasible solution to increase service capacity and convenience in a cost-efficient manner. It is much more likely that MPL will attract customer during the start-up phase due to their novelty in delivery mode. Therefore, having successfully diffused the MPL concept and got more customers on board, more stationary lockers (and thus cheaper) can be added to the existing distribution fleet without jeopardizing customer acceptance.

*Service radius and customer maximum walking range*

Empirical studies have revealed that customers will normally accept a walking distance of no more than 0.7 km for parcel collection (De Oliveira et al., 2017; Kedia et al., 2017). Recall that the result shown in Figure 16d, 17d, and 18d, the results suggest that the service radius of MPL $r_m$ should not exceed 5 km if there is a requirement to guarantee the cost-savings in different cost units over a maximum walking distance of customer $\rho_c$ not exceeding 0.7 km.

On the other hand, as we generate demand within a square area of 5km in length and width (see data generation code we provided), we determine that by reducing the service radius of each MPL $r_m$ by 1 km, the fleet size and total distance travelled will be reduced by at least 1 unit and 50 km, respectively. Therefore, the break-even point between the service radius of the MPL and the demand density is particularly pertinent for logistics operators. With this in mind, synergistic MPLs and other delivery methods (e.g., traditional van-based delivery, and stationary lockers) remain the high-potential efficient and cost-effective delivery scenarios in the area where demand density is high.

*5.5 Analysis and discussion*

In this section, we discuss the proposed method to deal with a special form of LRP, namely MPLP. The main difference between LRP and MPLP is that the movement trajectory of MPL will be determined based on all potential stopovers of the customer set while the classic LRP considers the static characteristics of the demand pattern.

The mathematical model is initially solved using an exact approach solver, Gurobi, to determine the gap between HQM and optimal solutions. The result (Table 5 and 6) shows that HQM has superior ability in deriving high-quality solutions and computational efficiency compared to the Gurobi solver for large problem instances, while it suffers from insufficient training samples for small-scale problems.

However, the exact approach has a weakness in that it takes a considerable amount of time to compute the result. In real-world scenarios, the magnitude of the problem often makes it impossible to solve the problem accurately within a reasonable timeframe (Yu et al., 2022a). Therefore, the results obtained from the exact approach are only used as a baseline for evaluating the solutions obtained by applying the proposed HQM.

Meta-heuristics, as a typical method used for solving the large problem instances of LRPs or VRPs, we, therefore, focus on the performance comparison between HQM and GA. The comparison covers two dimensions that provide a comprehensive overview of HQM's performance: i) solution quality and convergence/computation efficiency (e.g., Figure 12-15, Table 8); and ii) cost units and operation indications (e.g., Appendix D, Figure 11, Table 7). The results showed that HQM can provide high-quality solutions within a reasonable computational time when compared to GA solutions. However, there is no absolutely efficient algorithm, i.e., algorithms will suffer from the dilemma of compromising solution quality and compromising efficiency (Wolpert, 2013). Although the proposed HQM performs well in LRPs or VRPs and can provide a novel conceptual RL-based solution framework, further experiments are required to validate its generalisation ability to more complex scenarios.

There are several potential limitations of the proposed HQM model that may arise in larger problem instances. One limitation is that HQM may have differences in optimisation capabilities for different cost units within the objective function, even though the results of different cost units are influenced by the weights assigned to them. Nevertheless, information exchange and exploitation among different variables $x$ within a state $s$ will also have an indirect effect on the optimisation of different cost units since these variables serve as logical abstractions of various cost units. To address this limitation, it may be helpful to design an effective communication mechanism, such as the use of intermediate variables or transformer layers, between different variables to make the information interpretable and usable among the variables. Another limitation is the difficulty of interpreting the mathematical model into the RL framework, particularly when representing model constraints. One approach to address this is to decompose the original model such that the model constraints are processed and implemented outside the RL framework, while RL is only responsible for training the model. The last limitation is the computation time. Although HQM's computation efficiency is significantly superior to the exact approach, it still falls short of heuristic algorithms on large problem instances due to HQM's sampling procedure. The computational time required by HQM can be determined by analysing the procedure defined in the solution framework (Figure 5) in terms of complexity. The complexity of the proposed solution is in (1) the task generation procedure, (2) the reward calculation within the attention-based decoder, (3) the Q-value update procedure within the attention-based decoder, and (4) the global search and local search procedure. The other remaining modules can be executed in a constant time.

The procedure of task generation has a complexity of $O(|N \times K_n|)$ because we must enumerate all potential stopovers $K_n$ for customer set $N$ to identify a set of tasks that satisfy time-window constraints imposed by customers and parking spaces. The reward calculation within the attention-based decoder has the complexity of $O(|P|)$ since the decoder will calculate the corresponding reward for all state $s$ ($s \in \{1, \cdots, P\}$) within the agent $\pi$. In the case of Q-value updates, the computation complexity increases dramatically. Since there are two Q-value matrixes $Q\left(\theta_{x_1}^o\right)$ and $Q\left(\theta_{x_2}^o\right)$ (see Equation 29 and 30) in our case which corresponding to variable $x_1$ and $x_2$ of a state $s$ respectively, therefore HQM have to update the Q-value matrix simultaneously. Specifically, the complexity of updating $Q\left(\theta_{x_1}^o\right)$ is $O(M^2)$ where $M$ represents the total number of MPL, while the complexity of updating $Q\left(\theta_{x_2}^o\right)$ is $O(o^2)$ where $o$ represents the total number of tasks. When it comes to the global search procedure, since all elements within the state $s$ will be updated according to Q-value matrixes in each timestep, the complexity is $O(\Gamma)$, where $\Gamma$ represents the total number of elements $(\Gamma = |x_1 + x_2|)$ within the state $s$. Finally, the complexity of local search procedure is $O(\Upsilon)$, where $\Upsilon$ is the number of elements that need to be adjusted/shifted between two neighbourhood states. The complexities discussed above are the primary reason that the HQM's computation time increases with the sample size.

As alternatives to the encoder & attention-based decoder and heuristics introduced in our study, we provide the following methods to balance solution quality and computation efficiency. The first type of method focuses on generating superior initial solution and improving the efficiency of transcription of input samples. An example of the former method is introducing Negatively Corelated Search (NCS) during the training process to improve the diversity of better initial solutions (Tang et al., 2016). The latter method typically involves using batch reinforcement learning, which differs from traditional online learning that the learning agent does not interact directly with the environment. It uses a baseline agent that collects data, which is then fed into an algorithm to train a new policy (Laroche et al., 2019). Finally, an emerging technique is 'learning to delegate' approach, which improves the solution quality and computation efficiency by identifying appropriate subproblems and delegating their improvement to a black box sub-solver (S. Li et al., 2021). 'Learning to delegate' framework consists of two components: 1) a sub-solver capable of solving a small problem instance, and 2) a learning model for identifying suitable subproblems within a larger problem to delegate them to an exact solver.

The proposed mathematical model can be extended to solve other variant VRP-related problems. To adapt the mathematical model to a specific scenario, existing constraints must be modified (e.g., considering the parcel pick-up, demand heterogeneity). For the HQM, the encoder and attention-based decoder can be adopted to deal with other VRP variants with some modifications in the proposed solution interpretation, given the excellent generalisation ability of HQM's framework.

## 6 Conclusion

This study presents a novel approach for MPLP that accounts for changing customer demand patterns and varied scales, as well as stochastic events. A hybrid Q-Learning network algorithm was developed that combines global and local search mechanisms is presented, which resolve stochastic events that cause delays by employing two route adjusting strategies.

We first evaluate the performance of HQM by using an exact approach solver (Gurobi) as a benchmark to provide a first-hand insight of the solution quality and computation efficiency. The result shows that our HQM achieves higher solution quality with significantly shorter computation time than the Gurobi solver in large problem instances. To highlight the advantage of HQM compared with meta-heuristic algorithms that commonly used for solving LRPs and VRPs, we exploited a customised GA as baseline cases and conducting pairwise comparison under different network densities (15 to 200 nodes) with different route adjustment strategies applied. Our proposed HQM provides solutions in lower computing times than other comparable learning algorithms and obtains a final reward function twice as greater as our benchmark. Additionally, our result reveals that HQM achieves significant improvement in service delay and parcel first-time delivery successful rate compared to GA. Finally, the study investigates the influence of critical factors on different cost units within the objective function, providing logistics operators with parameterised and managerial insights for enhancing their delivery service.

In addition to the computational advantages of the HQM model, our results suggest that the use of HQM in combination with HCPS leads to reduced service delays and improved MPL configuration compared to using BTD. Although HQM shows a good optimisation ability in our MPLP scenario, further experiments and adaptions are required before migrating it to other LRP and VRP-related problems. To tackle the limitations of sampling efficiency and solution quality that may falls short compared with heuristics, we provide several potential solutions such as applying delegate agent, and batch reinforcement learning approaches (see Section 5.5).

We finally provide managerial insight based on the parametric analysis for different critical factors against cost units. The most critical factors that affect different cost units are the time window constraints imposed by parking spaces and customers, locker capacity, and network density. Logistics operators face the challenge of balancing deployment costs with improved delivery convenience and punctuality when broadly implementing MPLs in the market. Further, the result highlight that collaborating MPL with other mainstream delivery methods (e.g., traditional van-based delivery and stationary locker) offer the potential for broader application of MPL during start-up phase, especially in cities with high demand density and traffic congestion. As an alternative, dispatching MPLs before peak commuting hours based on customers' travel behaviour allows more demand to be fulfilled at more concentrated sites (e.g., company, school) and shorter time intervals of the day. From the perspective of urban planning and sustainability, deploying MPLs in less congested areas (e.g., residential areas) allows lockers to stay longer without disturbing traffic flow while the characteristic of their electric-powered mobility reduces gas emissions. However, more research is needed to examine the practicality and validity of integrating customers' daily locations with MPLs for last-mile deliveries, including the collection and protection of customer data.

Future work will focus on more complex dynamic cases of MPLP that require more flexible route adjustment approaches to respond to changes in demand and delays. It would also be useful to study multi-echelon MPLP, which involves the use of transhipment centres or feeder vehicles to reduce the distance between lockers and the central depot. Another extension of the problem is to consider cooperative deliveries in conjunction with MPLs, such as the use of couriers or delivery robots. An alternative approach is to segment the market so that the method of parcel delivery within a specific area is determined based on predetermined criteria such as demand density, road conditions, and traffic flow. From the perspective of algorithmic innovation, further research on advanced RL-based algorithms for solving MPLP and VRP is valuable. This could include implementing parallel, interoperable interaction between RL methods and the simulator to better match real-world operations. Finally, it would be useful to apply the model to real operational case studies in different scenarios, such as urban and rural regions, to validate the model.

**Appendix A.** Decoding algorithm for state

The input of the decoding algorithm is a state $s$ obtained from the encoder. The output is an interpretable task set of each MPL $m$. The mechanism of the decoding procedure is to traverse each element of $x_1$ within a state $s$, and assign it to the corresponding MPL following the sequence defined by $x_2$.

---

**Algorithm:** Decoding algorithm for route generation

---

Input: State $s = (x_1, x_2)$, Number of MPLs $M$

output: Task set $\bar{O}_m$ of MPLs $m$

1 *total_dispatch* $\leftarrow 0$ // Number of MPLs deployed

2 $\bar{O}_m \leftarrow \emptyset$

3 $x_1 \leftarrow s[0]$ // Variable of task assignment

4 $x_2 \leftarrow s[1]$ // Variable of task execution sequence

5 **for** $m$ *in* $M$ **do**

6      $\bar{O}_m[m] \leftarrow \emptyset$

7      *index_$x_1$* $\leftarrow$ argwhere $(x_1 == m)$ // Extract index where the tasks are allocated to MPL $m$

8      **if** *index_$x_1$* **is not** $\emptyset$ **then**

9          *index_$x_2$* $\leftarrow$ argwhere $(x_2 == index\_x_1)$

10          *index_$x_2$* $\leftarrow$ argsort $(index\_x_2)$ // Sorting index

11          *task* $\leftarrow$ *index_$x_1$[index_$x_2$]*

12          $\bar{O}_m[m] \leftarrow task$ // Task list of MPL $m$ with execution sequence

13          *total_dispatch* $+= 1$

14      **end if**

15 **end for**

16 **return** $\bar{O}_m$

---

**Appendix B.** Code listing of Genetic Algorithm (GA)

We design a customised GA for the comparison purpose with HQM in our study. The overall implementation process is as follows. We first create the population, in which the population size equals the number of states within the agent $\pi$ in HQM. The population is then ranked based on the fitness function (the value of Equation 3). Using an elite strategy and roulette wheel selection, individuals with higher fitness values will be selected to produce the next generation. The next step is to exploit the elite individuals to create a mating pool and breed the population. As a final step, we apply mutation operation to the new generation to obtain better solutions. Following the notation and parameter setting mentioned in Table 4, the code listing of GA is as follows:

---

**Algorithm:** Genetic algorithm

---

Input: Population size *Pop*; Possibility of crossover $P_{cross}$; Proportion of the elite size $P_{elite}$; Mutation rate $P_{muta}$; Number of MPLs *M*; All task list *O*

output: The optimal individual *individual*\*

1 *individual* $\leftarrow \emptyset$; *population* $\leftarrow \emptyset$; *fitness* $\leftarrow \emptyset$; *pop_index* $\leftarrow \emptyset$; *pop_ranked* $\leftarrow \emptyset$; *cum_perc* $\leftarrow \emptyset$; *mating_pool* $\leftarrow \emptyset$; *children* $\leftarrow \emptyset$

2 **for** *i* **in** range(*Pop*) **do** // Initialise population

3 $\quad$ $x_1 \leftarrow \mathbb{N} \sim U(0, \|M\|)$, $x_2 \leftarrow$ Permutating task list *O*

4 $\quad$ *individual* $= (x_1, x_2)$

5 $\quad$ Add *individual* to *population*

6 **end for**

5 **for** *individual* **in** *population* **do** // Rank population

6 $\quad$ *fitness* $\leftarrow$ Calculate the fitness value of *individual* according to **Equation (4)**

7 $\quad$ *pop_index* $\leftarrow$ Obtain the index of the sorted fitness score in *fitness*

8 $\quad$ *pop_ranked* $= [population[i]$ for $i$ in *pop_index*$]$

9 **end for**

10 *cum_perc* $\leftarrow$ Calculate the cumulative percentage of the fitness value in *pop_ranked*

11 **for** *i* **in** range (len(pop_ranked) - $P_{elite}$, len(pop_ranked)) **do** // Select elite individuals within population

12 $\quad$ Add $P_{elite}$ elite individuals to *mating_pool*

13 **end for**

14 **for** *i* **in** range (0, len(pop_ranked) - $P_{elite}$) **do** // Fill the rest of population via roulette wheel selection

15 $\quad$ $P_{cross} \leftarrow 100 * U(0, 1)$ //Generate random probability

16 $\quad$ **for** *fitness_score* **in** *cum_perc* **do**

17 $\quad$ $\quad$ **if** $P_{cross} \leq fitness\_score$ **then**

18 $\quad$ $\quad$ $\quad$ Add the corresponding individual to *mating_pool* // Create the mating poo;

19 $\quad$ $\quad$ **end if**

20 $\quad$ **end for**

21 **end for**

22 **for** *i* **in** range (0, $P_{elite}$) **do** // Produce the next generation

23 $\quad$ Add the top $P_{elite}$ elite individual to *children*

24 **end for**

25 **for** *i* **in** range (0, len(*mating_pool*)- $P_{elite}$) **do** // Regenerate the rest of individual of *mating_pool*

26 $\quad$ Exchange some chromosome of the two adjacent individuals // Breed the population

27 $\quad$ Obtain the updated population *population'*

28 **end for**

---

```
29 for individual in population′ do // Mutation operation
30 │   for chromosome in individual do
31 │   │   prob ← 100 * U(0, 1) // Generate random probability for mutation
32 │   │   if P_muta ≤ prob then
33 │   │   │   Exchange the current value of the chromosome with the adjacent chromosome
34 │   │   │   Obtain the updated population population″
34 │   │   end if
35 │   end for
36 end for
37 Recalculate and rank the fitness value of the new population population″
38 Output the best individual individual* with the highest fitness score if convergence condition satisfies
39 Return individual*
```

## Appendix C. Performance of the Gurobi solver and HQM under different problem instances

This appendix focuses on the implementation result comparison of Gurobi and HQM in terms of MPLs deployment size, total travel distance, and service delays. It should be noticed that the value of different cost units of HQM is the mean value between BTD and HCPS for the corresponding cost unit.

**Table C.1** MPLs deployed, travel distance, and service delay of Gurobi and HQM

| Number of parking spaces $I_i$ | Customer locations within each parking space $N_n$ | Gurobi | | | HQM | | | $Gap^a$ (%) | $Gap^b$ (%) | $Gap^c$ (%) |
|---|---|---|---|---|---|---|---|---|---|---|
| | | MPLs deployed (unit) | Travel distance (km) | Service delay (min) | MPLs deployed (unit) | Travel distance (km) | Service delay (min) | | | |
| 5 | 5 | 8 | 97.988 | 0 | 9 | 109.519 | 0 | 12.50 | 11.77 | 0 |
| | 10 | 12 | 177.869 | 0 | 15 | 183.984 | 0 | 20.83 | 3.44 | 0 |
| | 15 | 16 | 174.453 | 0 | 16 | 194.428 | 0.022 | -3.13 | 11.45 | 0 |
| | 20 | 16 | 240.409 | 0 | 16 | 221.891 | 7.180 | 0.00 | -7.70 | - |
| 6 | 5 | 9 | 128.190 | 0 | 11 | 139.608 | 0 | 22.22 | 8.91 | 0 |
| | 10 | 15 | 171.281 | 0.721 | 16 | 224.479 | 0.965 | 3.33 | 31.06 | 33.84 |
| | 15 | 18 | 254.115 | 5.684 | 18 | 264.484 | 2.583 | -2.78 | 4.08 | -54.56 |
| | 20 | 22 | 297.026 | 2.341 | 22 | 278.925 | 0.606 | -2.27 | -6.09 | -74.13 |
| 7 | 5 | 12 | 154.041 | 0 | 12 | 173.707 | 0 | 0.00 | 12.77 | 0 |
| | 10 | 19 | 244.694 | 5.076 | 20 | 278.558 | 11.076 | 2.63 | 13.84 | 118.20 |
| | 15 | 25 | 451.235 | 20.958 | 24 | 341.547 | 15.759 | -4.00 | -24.31 | -24.81 |
| | 20 | 28 | 408.578 | 25.618 | 26 | 423.333 | 13.996 | -8.93 | 3.61 | -45.37 |
| 8 | 5 | 15 | 171.184 | 0 | 15 | 186.306 | 0 | 0.00 | 8.83 | 0 |
| | 10 | 20 | 401.430 | 15.438 | 19 | 309.036 | 10.524 | -5.00 | -23.02 | -31.83 |
| | 15 | 25 | 566.556 | 14.359 | 23 | 371.916 | 21.252 | -10.00 | -34.35 | 48.00 |
| | 20 | 33 | 707.538 | 20.113 | 33 | 456.919 | 19.589 | -1.52 | -35.42 | -2.61 |
| 9 | 5 | 14 | 195.111 | 0 | 15 | 204.009 | 0.12 | 7.14 | 4.56 | - |
| | 10 | 22 | 348.313 | 14.293 | 22 | 383.603 | 23.471 | 0.00 | 10.13 | 64.21 |
| | 15 | 33 | 515.278 | 20.439 | 30 | 421.291 | 5.613 | -9.09 | -18.24 | -72.54 |
| | 20 | 33 | 575.891 | 19.872 | 31 | 470.912 | 19.691 | -6.06 | -18.23 | -0.91 |
| 10 | 5 | 18 | 222.569 | 0 | 19 | 252.073 | 0.606 | 2.78 | 13.26 | - |
| | 10 | 26 | 339.106 | 31.611 | 24 | 402.434 | 19.333 | -9.62 | 18.67 | -38.84 |
| | 15 | 34 | 519.177 | 19.243 | 31 | 476.686 | 19.437 | -8.82 | -8.18 | 1.01 |
| | 20 | 34 | 776.938 | 35.987 | 33 | 561.443 | 25.866 | -2.94 | -27.74 | -28.12 |

**Notes:** i) $Gap^a$ = (MPLs deployed of HQM - MPLs deployed of Gurobi)/ MPLs deployed of Gurobi; ii) $Gap^b$ =(Travel distance of HQM - Travel distance of Gurobi)/ Travel distance of Gurobi; iii) $Gap^c$ = (Service delay of HQM – Service delay of Gurobi)/ Service delay of Gurobi;

# Appendix D. Performance of HQM and GA under different problem instances

This appendix focuses on the implementation result comparison of HQM and GA in terms of MPLs deployment size and the total travel distance. The comparison is based on two perspectives: i) HQM vs. GA, and 2) BTD vs. HCPS.

**Table D.1** MPLs deployed and travel distance of HQM and GA

| Parking spaces $I_i$ | Customers per parking space $N_n$ | Route adjustment policy | HQM | | GA | | Gap (HQM vs. GA) | |
|---|---|---|---|---|---|---|---|---|
| | | | MPLs deployed (unit) | Travel distance (km) | MPLs deployed (unit) | Travel distance (km) | MPLs deployed (unit) | Travel distance (km) |
| 5 | 5 | BTD | 9 | 115.196 | 12 | 107.922 | -3 | 7.274 |
| | | HCPS | 9 | 103.842 | 15 | 125.609 | -6 | -21.767 |
| | | Gap (BTD vs. HCPS) | 0 | 11.354 | -3 | -17.687 | - | - |
| | 10 | BTD | 13 | 188.929 | 16 | 168.477 | -3 | 20.452 |
| | | HCPS | 16 | 179.039 | 16 | 175.058 | 0 | 3.981 |
| | | Gap (BTD vs. HCPS) | -3 | 9.89 | 0 | -6.581 | - | - |
| | 15 | BTD | 15 | 203.999 | 16 | 178.342 | -1 | 25.657 |
| | | HCPS | 16 | 184.857 | 14 | 162.803 | 2 | 22.054 |
| | | Gap (BTD vs. HCPS) | -1 | 19.142 | 2 | 15.539 | - | - |
| | 20 | BTD | 17 | 230.168 | 16 | 199.032 | 1 | 31.136 |
| | | HCPS | 15 | 213.614 | 15 | 216.142 | 0 | -2.528 |
| | | Gap (BTD vs. HCPS) | 2 | 16.554 | 1 | -17.11 | - | - |
| 6 | 5 | BTD | 11 | 148.131 | 11 | 135.441 | 0 | 12.69 |
| | | HCPS | 11 | 131.085 | 11 | 127.910 | 0 | 3.175 |
| | | Gap (BTD vs. HCPS) | 0 | 17.046 | 0 | 7.531 | - | - |
| | 10 | BTD | 15 | 214.336 | 17 | 205.476 | -2 | 8.86 |
| | | HCPS | 16 | 234.621 | 14 | 196.877 | 2 | 37.744 |
| | | Gap (BTD vs. HCPS) | -1 | -20.285 | 3 | 8.599 | - | - |
| | 15 | BTD | 18 | 268.043 | 17 | 227.272 | 1 | 40.771 |
| | | HCPS | 17 | 260.924 | 17 | 214.830 | 0 | 46.094 |
| | | Gap (BTD vs. HCPS) | 1 | 7.119 | 0 | 12.442 | - | - |
| | 20 | BTD | 23 | 274.511 | 21 | 277.209 | 2 | -2.698 |
| | | HCPS | 20 | 283.338 | 21 | 261.223 | -1 | 22.115 |
| | | Gap (BTD vs. HCPS) | 3 | -8.827 | 0 | 15.986 | - | - |
| 7 | 5 | BTD | 12 | 179.269 | 12 | 157.566 | 0 | 21.703 |
| | | HCPS | 12 | 168.145 | 15 | 170.794 | -3 | -2.649 |
| | | Gap (BTD vs. HCPS) | 0 | 11.124 | -3 | -13.228 | - | - |
| | 10 | BTD | 20 | 271.938 | 19 | 236.719 | 1 | 35.219 |
| | | HCPS | 19 | 285.177 | 18 | 224.509 | 1 | 60.668 |
| | | Gap (BTD vs. HCPS) | 1 | -13.239 | 1 | 12.21 | - | - |
| | 15 | BTD | 23 | 345.110 | 23 | 281.598 | 0 | 63.512 |
| | | HCPS | 25 | 337.984 | 23 | 279.518 | 2 | 58.466 |
| | | Gap (BTD vs. HCPS) | -2 | 7.126 | 0 | 2.08 | - | - |
| | 20 | BTD | 23 | 371.527 | 25 | 346.943 | -2 | 24.584 |
| | | HCPS | 28 | 475.139 | 24 | 373.877 | 4 | 101.262 |
| | | Gap (BTD vs. HCPS) | -5 | -103.612 | 1 | -26.934 | - | - |
| 8 | 5 | BTD | 15 | 188.846 | 16 | 163.958 | -1 | 24.888 |
| | | HCPS | 15 | 183.766 | 15 | 174.367 | 0 | 9.399 |
| | | Gap (BTD vs. HCPS) | 0 | 5.08 | 1 | -10.409 | - | - |
| | 10 | BTD | 19 | 315.499 | 19 | 248.248 | 0 | 67.251 |
| | | HCPS | 19 | 302.573 | 19 | 251.514 | 0 | 51.059 |
| | | Gap (BTD vs. HCPS) | 0 | 12.926 | 0 | -3.266 | - | - |
| | 15 | BTD | 22 | 379.610 | 24 | 361.005 | -2 | 18.605 |
| | | HCPS | 23 | 364.222 | 23 | 333.354 | 0 | 30.868 |
| | | Gap (BTD vs. HCPS) | -1 | 15.388 | 1 | 27.651 | - | - |
| | 20 | BTD | 32 | 459.328 | 30 | 414.339 | 2 | 44.989 |
| | | HCPS | 33 | 454.509 | 31 | 406.337 | 2 | 48.172 |
| | | Gap (BTD vs. HCPS) | -1 | 4.819 | -1 | 8.002 | - | - |
| 9 | 5 | BTD | 15 | 202.043 | 15 | 192.197 | 0 | 9.846 |
| | | HCPS | 15 | 205.974 | 16 | 194.847 | -1 | 11.127 |
| | | Gap (BTD vs. HCPS) | 0 | -3.931 | -1 | -2.65 | - | - |
| | 10 | BTD | 22 | 378.835 | 22 | 360.514 | 0 | 18.321 |
| | | HCPS | 22 | 388.371 | 21 | 329.978 | 1 | 58.393 |
| | | Gap (BTD vs. HCPS) | 0 | -9.536 | 1 | 30.536 | - | - |
| | 15 | BTD | 29 | 432.421 | 30 | 411.984 | -1 | 20.437 |
| | | HCPS | 31 | 410.161 | 28 | 387.925 | 3 | 22.236 |
| | | Gap (BTD vs. HCPS) | -2 | 22.26 | 2 | 24.059 | - | - |
| | 20 | BTD | 32 | 480.981 | 32 | 450.828 | 0 | 30.153 |
| | | HCPS | 30 | 460.842 | 30 | 419.316 | 0 | 41.526 |
| | | Gap (BTD vs. HCPS) | 2 | 20.139 | 2 | 31.512 | - | - |
| 10 | 5 | BTD | 18 | 254.706 | 16 | 204.258 | 2 | 50.448 |
| | | HCPS | 19 | 249.439 | 16 | 217.141 | 3 | 32.298 |
| | | Gap (BTD vs. HCPS) | -1 | 5.267 | 0 | -12.883 | - | - |
| | 10 | BTD | 24 | 420.902 | 24 | 345.720 | 0 | 75.182 |
| | | HCPS | 23 | 383.965 | 23 | 335.836 | 0 | 48.129 |
| | | Gap (BTD vs. HCPS) | 1 | 36.937 | 1 | 9.884 | - | - |
| | 15 | BTD | 31 | 491.675 | 29 | 435.426 | 2 | 56.249 |
| | | HCPS | 31 | 461.697 | 30 | 446.390 | 1 | 15.307 |
| | | Gap (BTD vs. HCPS) | 0 | 29.978 | -1 | -10.964 | - | - |
| | 20 | BTD | 34 | 580.228 | 31 | 507.270 | 3 | 72.958 |
| | | HCPS | 32 | 542.657 | 32 | 515.494 | 0 | 27.163 |
| | | Gap (BTD vs. HCPS) | 2 | 37.571 | -1 | -8.224 | - | - |

**Notes:** 1) **Gap (HQM vs. GA)** = HQM – GA; 2) **Gap (BTD vs. HCPS)** = BTD - HCPS